\documentclass{bmvc2k}

\usepackage{amsmath,amssymb}
\usepackage{pdfpages}

\AtBeginDocument{%
  \pdfpagewidth=456pt %
  \hoffset=32pt %
}
\usepackage{booktabs}
\usepackage{multirow}
\usepackage{array}

\graphicspath{{figures/}}

\providecommand{\keywords}[1]{}

\begin{document}

\title{LinearMask-GS: Stable-Mask Importance Pruning for
       Compact 3D Gaussian Splatting}

\addauthor{Donghun Ryu}{donghun0621@cau.ac.kr}{1}
\addauthor{Minhyeok Lee}{mlee@cau.ac.kr}{1,2$\dagger$}
\addinstitution{%
  Department of Intelligent Semiconductor Engineering\\
  Chung-Ang University\\
  Seoul, Republic of Korea}
\addinstitution{%
  School of Electrical and Electronics Engineering\\
  Chung-Ang University\\
  Seoul, Republic of Korea}
\runninghead{Ryu, Lee}{LinearMask-GS}

\maketitle
\makeatletter
\BMVA@blfootnote{\null\hspace{-1.9em}$^{\dagger}$Corresponding author.}
\BMVA@blfootnote{\null\hspace{-1.9em}Accepted to the British Machine Vision Conference (BMVC) 2026.}
\makeatother
\setcounter{footnote}{0}

\begin{abstract}
3D Gaussian Splatting~(3DGS) enables real-time novel view synthesis but
produces millions of primitives through adaptive densification,
leading to significant storage overhead.
Learned-mask pruning methods such as LP-3DGS address this by assigning
each Gaussian a learnable mask to identify and prune redundant primitives.
However, we identify a limitation of this paradigm: the steep slope of the Gumbel-Sigmoid activation drives mask values to the extremes within the short mask-training window, before the importance ranking has stabilized, producing a sharply bimodal distribution from which that ranking can no longer be reliably recovered.
We propose LinearMask-GS, which replaces Gumbel-Sigmoid with a linear increment activation that keeps mask values in a mid-confidence regime throughout mask training, producing a stable, unimodal mask distribution whose ranking tracks importance.
On Mip-NeRF 360, our method achieves $3.6\times$ and $1.6\times$ Gaussian reductions over 3DGS and LP-3DGS, respectively, while maintaining or improving rendering quality. For outdoor scenes, it yields a $1.6\times$ reduction (from 2.18M to 1.36M) with notable gains in PSNR (+0.38 dB), SSIM (+0.025), and LPIPS (-0.029).
\end{abstract}

\section{Introduction}
\label{sec:introduction}

3D Gaussian Splatting (3DGS)~\cite{kerbl20233d} has become a leading
representation for novel view synthesis, achieving high-fidelity real-time
rendering by encoding scenes as millions of explicit Gaussian primitives.
Despite this rendering efficiency, the adaptive densification process
routinely produces millions of primitives, and a standard outdoor scene in
Mip-NeRF~360~\cite{barron2022mipnerf360} can exceed 1GB in
storage~\cite{xie2024mesongs}, limiting deployment on
resource-constrained devices.

Prior methods reduce this overhead either by compressing per-Gaussian
attributes~\cite{niedermayr2024compressed3dgs, lee2024compact3dgs,
xie2024mesongs}, which leaves the primitive count and rasterization cost
unchanged, or by pruning low-importance Gaussians post
hoc~\cite{fan2023lightgaussian, fang2024minisplatting,
niemeyer2024radsplat}, which lacks feedback from scene optimization and
often discards geometrically essential primitives, yielding suboptimal
quality--compression trade-offs (Sec.~\ref{sec:related_work}).

LP-3DGS~\cite{kim2024lp3dgs} addresses this by learning a per-Gaussian soft
mask jointly with the scene, using a Gumbel-Sigmoid activation so the mask
can be optimized end-to-end and then thresholded into a hard keep/prune
decision. This couples pruning with scene optimization, but we observe that
the soft mask it produces is, by the time pruning happens, no longer a
faithful summary of importance.

\textbf{The learned mask saturates before importance ranking stabilizes.}
The Gumbel-Sigmoid activation used in LP-3DGS is steep: small changes in the
underlying parameter swing the mask all the way to $0$ or $1$, so the mask
collapses to these two extreme values long before the per-Gaussian
importance signal has settled into a meaningful ranking.
This saturation is a property of the activation itself rather than of the
brevity of the mask-training window: with the pruning point held fixed,
extending the window monotonically \emph{worsens} saturation
(Sec.~\ref{subsec:linear_mask}).
As shown in Fig.~\ref{fig:mask_histogram}(a, b), the resulting distribution
is sharply two-peaked (i.e., bimodal), with almost every value pinned
to either $0$ or $1$, and which side a Gaussian lands on is largely an
artifact of early optimization rather than its true contribution to the
rendered image.
Pruning a distribution like this is unreliable: the ranking induced by
$\psi_i$ reflects the side of saturation more than the magnitude of
importance, and the standard remedy of enforcing a minimum-survival
ratio~\cite{kim2024lp3dgs} prevents geometry collapse without addressing
this underlying cause.

To address this, we propose \textbf{LinearMask-GS}, a learned-mask pruning
framework that targets the cause of this brittleness by changing how the
mask is shaped during training.
We replace Gumbel-Sigmoid with a linear increment activation whose
gentle slope keeps mask values in the middle of the $[0,1]$ range, a
mid-confidence regime, throughout the mask-training window, rather
than letting them saturate at $0$ or $1$.
The resulting distribution (Fig.~\ref{fig:mask_histogram}(c, d)) is
unimodal, so the relative order of mask values is preserved rather than collapsed onto $0$ and $1$, and hard pruning therefore operates on a stable ranking rather than a two-peaked one.
Because the activation only changes how the mask is shaped, not how it is
applied, LinearMask-GS integrates without architectural modifications into a range of Gaussian-based backbones, including 3DGS~\cite{kerbl20233d}, 2DGS~\cite{huang20242d}, DropGaussian~\cite{park2025dropgaussian}, Octree-GS~\cite{ren2024octreegs}, and FastGS~\cite{ren2025fastgs}.

Our main contributions are as follows:
\vspace{-\topsep}
\begin{itemize}
    \item We identify that the steep activation of learned-mask pruning
    saturates the soft mask at $0$ or $1$ before the importance ranking
    stabilizes, a failure mode that longer mask-training windows worsen
    rather than fix, yielding a two-peaked distribution from which that
    ranking can no longer be reliably recovered.
    \item We propose the linear increment masking activation, whose
    gentle slope keeps mask values away from the boundaries and yields a
    unimodal distribution whose ranking is preserved rather than collapsed, on which a simple top-$\rho N$
    selection gives a reliable hard-pruning decision.
    \item On Mip-NeRF~360, Tanks~\&~Temples, and Deep~Blending,
    LinearMask-GS achieves better quality--compression trade-offs than
    existing learned-mask and pruning baselines on 3DGS, and the same
    masking module transfers without modification to 2DGS, DropGaussian,
    Octree-GS, and FastGS.
\end{itemize}

\section{Related Work}
\label{sec:related_work}

\subsection{Novel View Synthesis}
While Neural Radiance Fields (NeRF)~\cite{mildenhall2021nerf, barron2022mipnerf360, muller2022instant} achieve high-quality novel view synthesis, dense ray marching limits their real-time application. 3D Gaussian Splatting (3DGS)~\cite{kerbl20233d} overcomes this by combining explicit 3D Gaussian primitives with a fast tile-based rasterizer. Subsequent methods have extended this framework to improve anti-aliasing~\cite{yu2024mipsplatting}, scaling~\cite{lu2024scaffold}, and surface reconstruction via 2D oriented disks (2DGS)~\cite{huang20242d}. 
Despite these advances, adaptive densification leaves all of them with millions of primitives and the attendant memory and storage overhead (Sec.~\ref{sec:introduction}). 

\subsection{Gaussian Compression and Pruning}
To reduce storage costs without modifying the number of Gaussians, several works compress per-Gaussian attributes via vector quantization, residual quantization of color and rotation, or post-training codecs based on entropy coding and region-adaptive hierarchical transforms~\cite{niedermayr2024compressed3dgs, girish2024eagles, lee2024compact3dgs, xie2024mesongs}.
While effective at reducing per-attribute storage, these methods leave the total number of Gaussian primitives unchanged and thus do not alleviate rasterization time complexity.

Pruning methods directly reduce the Gaussian count by removing redundant
primitives.
The original 3DGS~\cite{kerbl20233d} periodically culls Gaussians with low
opacity or excessively large scale as part of adaptive density control.
LightGaussian~\cite{fan2023lightgaussian} prunes low-scoring primitives by a heuristic significance score, and Mini-Splatting~\cite{fang2024minisplatting} reorganizes Gaussian positions via a densification-and-sampling pipeline.
RadSplat~\cite{niemeyer2024radsplat} shows that taking the per-view maximum
blending weight as the importance score yields a more stable and
discriminative signal than summation-based alternatives.
Despite their utility, these heuristic methods rely on manually tuned thresholds and treat pruning as a post-hoc step decoupled from scene geometry optimization, leading to accidental removal of geometrically essential primitives.
Our method overcomes these limitations by coupling pruning decisions directly with scene reconstruction through learned importance masks.

\subsection{Differentiable Gaussian Pruning}
To overcome the limitations of manual thresholding, recent work introduces end-to-end differentiable masking frameworks.
LP-3DGS~\cite{kim2024lp3dgs} assigns a learnable binary mask to each Gaussian and jointly optimizes masks alongside scene parameters, accumulating importance scores by summing pixel-level blending weights and binarizing them through a Gumbel-Sigmoid activation.
Taming-3DGS~\cite{papantonakis2024taming} and Reduced-3DGS~\cite{papantonakis2024reduced} extend this paradigm with explicit Gaussian count budgets and attribute quantization, respectively, targeting deployment on resource-limited hardware.

While these methods demonstrate the benefit of jointly learning pruning decisions and scene geometry, they inherit summation-based score inflation and, more fundamentally, the premature saturation of the steep Gumbel-Sigmoid mask; our approach addresses both with max-pooling aggregation and a linear-increment activation that restores a stable ranking for the top-$\rho N$ rule (Secs.~\ref{subsec:importance_score}--\ref{subsec:linear_mask}).

\section{Method}
\label{sec:method}

We propose a differentiable pruning framework that jointly optimizes per-Gaussian mask parameters alongside scene geometry, compressing Gaussian-based representations while largely preserving PSNR, SSIM, and LPIPS. We develop it on 3DGS~\cite{kerbl20233d} and 2DGS~\cite{huang20242d}, and show that it transfers without modification to other backbones (Sec.~\ref{sec:experiments}).

After an initial geometric warm-up phase using the standard Gaussian Splatting pipeline, the framework proceeds in three stages:
(1) computing a max-pooled importance score $S_i$ for each primitive to evaluate view-count-invariant relevance (Sec.~\ref{subsec:importance_score}),
(2) jointly optimizing per-Gaussian learnable mask parameters with the proposed linear increment activation, which keeps mask values away from $0$ and $1$ throughout the mask-training window (Sec.~\ref{subsec:linear_mask}), and
(3) applying a hard-pruning rule that retains the top-$\rho N$ Gaussians ranked by $\psi_i$, where $\rho$ is the minimum-survival ratio, followed by fine-tuning of the remaining scene (Sec.~\ref{subsec:training_objective}).

\subsection{Preliminaries}
\label{subsec:prelim}

\paragraph{3D Gaussian Splatting.}
3DGS~\cite{kerbl20233d} represents a scene as a set of $N$ anisotropic Gaussian
primitives.
Each Gaussian $i$ is parameterized by a mean position $\boldsymbol{\mu}_i \in
\mathbb{R}^3$, a covariance matrix $\boldsymbol{\Sigma}_i = \mathbf{R}_i
\mathbf{S}_i \mathbf{S}_i^\top \mathbf{R}_i^\top$ (decomposed into a rotation
matrix $\mathbf{R}_i$ and a diagonal scaling matrix $\mathbf{S}_i$), an opacity
logit $o_i \in \mathbb{R}$, and spherical harmonic coefficients encoding
view-dependent color $\mathbf{c}_i$.

Primitives are projected and depth-sorted to compute color $\mathbf{C}$ via front-to-back alpha compositing:
\begin{equation}
    \mathbf{C} = \sum_{i=1}^{N} \mathbf{c}_i \, \alpha_i \prod_{j < i}
    (1 - \alpha_j), \quad
    \alpha_i = \sigma(o_i) \cdot \exp\!\left(
    -\tfrac{1}{2} \mathbf{d}^\top \boldsymbol{\Sigma}^{-1}_{i,\text{2D}}
    \mathbf{d} \right),
    \label{eq:rendering}
\end{equation}
Here, $\mathbf{d}$ is the 2D offset to the projected center, $\boldsymbol{\Sigma}_{i,\text{2D}}$ is the covariance, and $\sigma(\cdot)$ is the sigmoid.
The per-Gaussian blending weight at pixel $\mathbf{p}$ under viewpoint $k$ is:
\begin{equation}
    w_i^{k,\mathbf{p}} = \alpha_i^{k,\mathbf{p}}
    \prod_{j < i} (1 - \alpha_j^{k,\mathbf{p}}),
    \label{eq:blending_weight}
\end{equation}
which measures the fractional contribution of Gaussian $i$ to the rendered
pixel color.

\paragraph{2D Gaussian Splatting.}
2DGS~\cite{huang20242d} replaces 3D ellipsoids with oriented 2D disk
primitives embedded in 3D space, improving multi-view geometric consistency.
Its compositing follows the same front-to-back formulation as
Eq.~\eqref{eq:rendering} (with an explicit ray-splat intersection), and the
blending weight $w_i^{k,\mathbf{p}}$ retains the form of
Eq.~\eqref{eq:blending_weight}, providing a unified interface for importance
score computation across both representations.

\subsection{Importance Score via Max-Pooling}
\label{subsec:importance_score}

A key design choice in differentiable Gaussian pruning is how to aggregate
per-pixel, per-view blending weights into a single scalar importance score
$S_i$ for each Gaussian.
Several prior methods adopt summation-based aggregation~\cite{fang2024minisplatting,
kim2024lp3dgs}:
\begin{equation}
    S_i^{\text{sum}} = \sum_{k=1}^{K} \sum_{\mathbf{p}} w_i^{k,\mathbf{p}},
    \label{eq:sum_score}
\end{equation}
where the sum runs over all $K$ training viewpoints and all pixels $\mathbf{p}$.
This formulation inflates scores for Gaussians in densely observed regions,
reducing the discriminability of the importance signal.

To address this, we adopt the max-pooling aggregation proposed by RadSplat~\cite{niemeyer2024radsplat} to compute importance scores over all training views:
\begin{equation}
    S_i = \max_{k \in \{1, \ldots, K\}} \Bigl( \max_{\mathbf{p}} w_i^{k,\mathbf{p}} \Bigr).
    \label{eq:max_score}
\end{equation}
This captures the peak contribution of a Gaussian in its most favorable
viewpoint, providing a view-count-invariant measure of geometric relevance.
A Gaussian that contributes strongly to at least one view retains a high score
regardless of how many views overlap the same region, effectively mitigating the
view-count inflation associated with Eq.~\eqref{eq:sum_score}.

\subsection{Differentiable Pruning via Linear Increment Masking}
\label{subsec:linear_mask}

\paragraph{Mask formulation.}
We assign each Gaussian $i$ a learnable scalar parameter $m_i \in \mathbb{R}$,
initialized to~1 to ensure all primitives start fully visible and participate equally in the initial geometric optimization.
The per-Gaussian mask value is:
\begin{equation}
  \psi_i = f(m_i \cdot S_i),
  \label{eq:mask_value}
\end{equation}
where $f : \mathbb{R} \to [0,1]$ is the mask activation and $S_i$ is the
importance score from Eq.~\eqref{eq:max_score}.
During rendering, the opacity of each Gaussian is modulated as:
\begin{equation}
  \tilde{\alpha}_i^{k,\mathbf{p}} = \psi_i \cdot \alpha_i^{k,\mathbf{p}},
  \label{eq:mask_modulation}
\end{equation}
so that Gaussians with $\psi_i \approx 0$ are differentiably suppressed
while gradients flow through both $m_i$ and the scene parameters
concurrently.

\begin{figure}[!ht]
  \centering
  \begin{minipage}[b]{0.48\linewidth}
    \centering
    
    \includegraphics[width=\linewidth]%
      {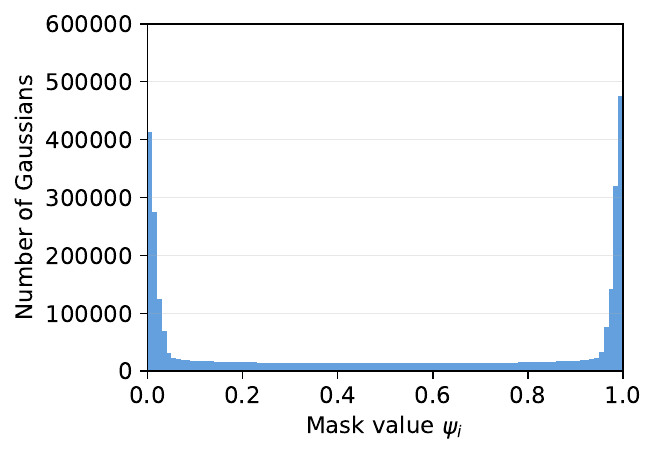}
    \centerline{\small (a) Gumbel-Sigmoid, outdoor scenes}
    \label{fig:mask_hist_gumbel_outdoor}
  \end{minipage}
  \hfill
  \begin{minipage}[b]{0.48\linewidth}
    \centering
    
    \includegraphics[width=\linewidth]%
      {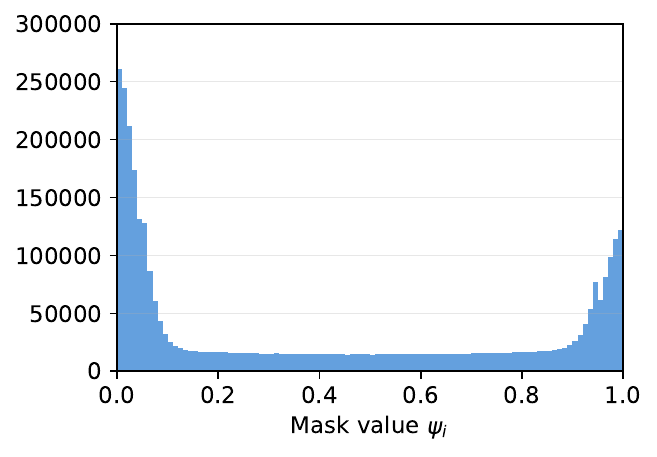}
    \centerline{\small (b) Gumbel-Sigmoid, indoor scenes}
    \label{fig:mask_hist_gumbel_indoor}
  \end{minipage}
  \vspace{0.4em}

  \begin{minipage}[b]{0.48\linewidth}
    \centering
    \includegraphics[width=\linewidth]%
      {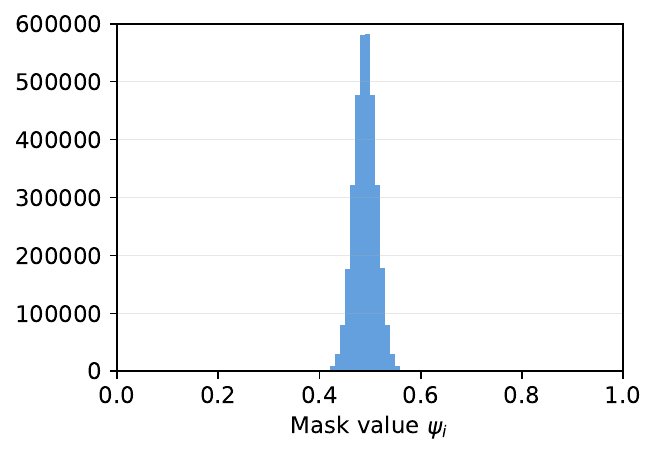}
    \centerline{\small (c) Linear increment (Ours), outdoor scenes}
    \label{fig:mask_hist_linear_outdoor}
  \end{minipage}
  \hfill
  \begin{minipage}[b]{0.48\linewidth}
    \centering

    \includegraphics[width=\linewidth]%
      {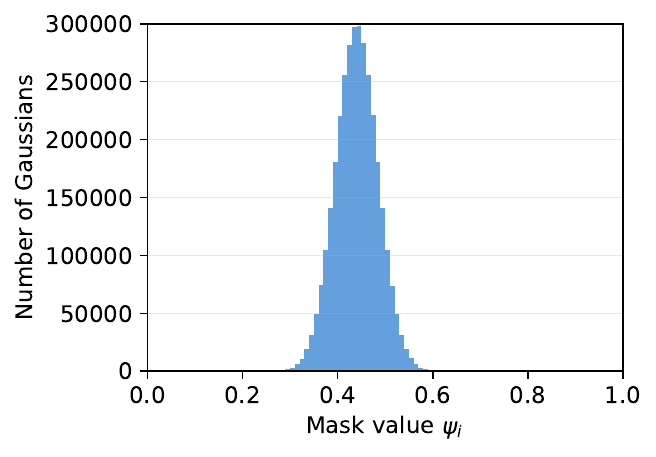}
    \centerline{\small (d) Linear increment (Ours), indoor scenes}
    \label{fig:mask_hist_linear_indoor}
  \end{minipage}
  \caption{%
    Mask $\psi_i$ distribution at the end of mask training, aggregated separately over Mip-NeRF~360's outdoor (5 scenes) and indoor (4 scenes) sets. (a, b) Gumbel-Sigmoid produces a two-peaked distribution pinned at $0$ and $1$; the fraction above $\psi=0.5$ is scene-dependent ($\sim$52\% outdoor, $\sim$40\% indoor), reflecting that which side each mask lands on is unrelated to importance. (c, d) Linear increment yields a unimodal distribution near $\psi=0.5$ (mean $\sim$0.44 indoor, $\sim$0.49 outdoor), so the relative order of mask values is preserved and the top-$\rho N$ pruning in Eq.~\eqref{eq:hard_pruning} operates on a stable ranking.
  }
  \label{fig:mask_histogram}
\end{figure}

\paragraph{Linear increment activation.}
LP-3DGS~\cite{kim2024lp3dgs} employs Gumbel-Sigmoid as $f$:
\begin{equation}
  f_{\text{GS}}(x)
    = \frac{1}{1 + \exp\!\left(-\frac{\log(x) + g_0 - g_1}{t}\right)},
    \quad g_0, g_1 \sim \mathrm{Gumbel}(0,1),
  \label{eq:gumbel_sigmoid}
\end{equation}
where $t$ is the temperature parameter.
The steep slope of $f_{\text{GS}}$ near its midpoint causes a small change in $m_i \cdot S_i$ to produce a large change in $\psi_i$, so mask outputs are pushed toward $0$ or $1$ within the short mask-training window before the importance score $S_i$ has stabilized.
As shown in Fig.~\ref{fig:mask_histogram}(a, b), this yields a two-peaked $\psi_i$ distribution pinned at $0$ and $1$. We refer to this behavior as mask instability: which extreme each mask lands on reflects an artifact of early optimization more than the magnitude of $S_i$ itself.
This instability stems from the activation's steepness rather than from the
window length: with the pruning point fixed at iteration 20k, extending the
Gumbel-Sigmoid mask window from $500$ to $2{,}500$ and $5{,}000$ iterations 
monotonically \emph{increases} the saturated fraction
($51.1\% \to 54.8\% \to 56.2\%$ on three scenes) while PSNR plateaus,
whereas the linear ranking is robust to window placement (window starts of
15k/17.5k/19.5k change final PSNR by $\leq 0.14$~dB); see Sec.~4 of the
supplementary material.

We therefore propose the linear increment activation:
\begin{equation}
  f_{\text{LI}}(x;\,\tau)
    = \mathrm{clip}\!\left(\tau \cdot (x - 0.5) + 0.5,\; 0,\; 1\right),
  \label{eq:linear_increment}
\end{equation}
where $\tau > 0$ is a slope hyperparameter. With a small slope (we use $\tau{=}0.1$), $f_{\text{LI}}$ has a wide unsaturated range and is monotone-linear in its active interval, so a change in $m_i \cdot S_i$ produces a proportionally small change in $\psi_i$. As a result, mask values stay close to the neutral midpoint $0.5$ throughout mask training and the output distribution remains unimodal and centered away from $0$ and $1$, as shown in Fig.~\ref{fig:mask_histogram}(c, d). We refer to this property as mask stability: because the values stay spread across the mid-range instead of collapsing to a binary indicator, their relative order is preserved rather than compressed onto the saturation points, so $\psi_i$ yields a stable ranking for pruning. Quantitatively, across the nine Mip-NeRF~360 scenes at iteration $20{,}000$, the Linear mask distribution shows $0.00\%$ saturation ($\psi_i < 0.1$ or $\psi_i > 0.9$) and $99.99\%$ of values lying in the interval $[0.3, 0.7]$, with a per-Gaussian standard deviation of only $0.018$. In contrast, Gumbel-Sigmoid saturates $51.19\%$ of mask values and retains only $20.28\%$ in $[0.3, 0.7]$, with a much larger standard deviation of $0.380$.

\paragraph{Hard pruning via top-$\rho N$ selection.}
At the end of mask training, we convert the soft mask values into a hard keep/prune decision. Because the linear-increment activation keeps $\psi_i$ unimodal and centered around $0.5$, the relative order of mask values is preserved rather than compressed onto the saturation points, so $\psi_i$ provides a stable ranking. We exploit this by directly retaining the top-$\rho N$ Gaussians by $\psi_i$, where $\rho = 0.30$ is the minimum-survival ratio and $N$ the total number of Gaussians:
\begin{equation}
  \mathcal{K} = \text{top-}\rho N \text{ Gaussians ranked by } \psi_i.
  \label{eq:hard_pruning}
\end{equation}
Gaussians in $\mathcal{K}$ are retained and the rest are permanently pruned. This rule plays the same role as the minimum-survival ratio in LP-3DGS (bounding the kept fraction to prevent geometry collapse), but because $\psi_i$ here is unimodal and its rank tracks importance, the kept set follows the underlying importance ranking instead of the early saturation outcome. 

Throughout the mask-training phase, the soft mask value $\psi_i \in [0,1]$ produced
by the linear-increment activation is applied directly as an opacity
multiplier via Eq.~\eqref{eq:mask_modulation}, so gradients propagate
through both $m_i$ and the scene parameters without hard discretization. The hard-pruning rule of Eq.~\eqref{eq:hard_pruning} is applied only once at the end of mask training, as described in Sec.~\ref{subsec:training_objective}.

\subsection{Training Objective and Schedule}
\label{subsec:training_objective}

The total training loss combines photometric reconstruction and mask sparsity:
\begin{equation}
    \mathcal{L}_{\text{total}} = (1 - \lambda_D)\,\mathcal{L}_1
    + \lambda_D\,(1 - \mathrm{SSIM})
    + \lambda_m \mathcal{L}_{\text{mask}},
    \label{eq:training_loss}
\end{equation}
where $\mathcal{L}_1$ is the pixel-wise $\ell_1$ loss, SSIM is the structural
similarity metric~\cite{wang2004image}, and $\lambda_D = 0.2$ follows the
default 3DGS setting~\cite{kerbl20233d}.
The sparsity term follows LP-3DGS~\cite{kim2024lp3dgs}:
\begin{equation}
  \mathcal{L}_{\text{mask}} = \frac{1}{N} \sum_{i=1}^{N} \lvert m_i \rvert,
  \label{eq:mask_loss}
\end{equation}
which encourages the mask parameters $m_i$ to shrink during training, thereby reducing the corresponding soft mask values after scaling by $S_i$.
We set $\lambda_m = 5 \times 10^{-4}$ and $\tau = 0.1$ for the
linear increment activation, using a separate AdamW optimizer~\cite{loshchilov2018decoupled} for $\{m_i\}$ at a learning rate of $5 \times 10^{-2}$.

\paragraph{Optimization schedule.}
Training proceeds for $30{,}000$ iterations in total.
For both the 3DGS and 2DGS backbones, we warm up the standard pipeline for $19{,}500$ iterations to ensure
geometric convergence before activating the mask module.
During the subsequent $500$-iteration mask training phase
(iterations $19{,}500$--$20{,}000$), the max-pooling importance scores $S_i$
from Eq.~\eqref{eq:max_score} are recomputed every $20$ iterations to reflect
ongoing geometric changes.
At iteration $20{,}000$, we apply the hard-pruning rule of Eq.~\eqref{eq:hard_pruning} introduced in Sec.~\ref{subsec:linear_mask}: the top-$\rho N$ Gaussians ranked by $\psi_i$ are kept, where $\rho$ is the minimum-survival ratio. We use $\rho = 0.30$ as the default and study its effect in Sec.~\ref{sec:experiments}.
The remaining primitives are fine-tuned without the mask loss for the
remainder of training.

\section{Experiments}
\label{sec:experiments}

\subsection{Experimental Setup}
\label{subsec:setup}

\paragraph{Datasets.}
We evaluate LinearMask-GS on three novel view synthesis benchmarks: Mip-NeRF~360~\cite{barron2022mipnerf360} (9~scenes), Tanks and Temples~\cite{knapitsch2017tanks} (2~outdoor scenes), and Deep Blending~\cite{hedman2018deepblending} (2~indoor scenes).

\paragraph{Baselines.}
We compare LinearMask-GS against generic novel view synthesis and scene representation baselines (Plenoxels~\cite{fridovich2022plenoxels}, INGP~\cite{muller2022instant}, Mip-NeRF~360~\cite{barron2022mipnerf360}, 3DGS~\cite{kerbl20233d}, and 2DGS~\cite{huang20242d}), as well as 3DGS-based pruning and compression methods (Compact3DGS~\cite{lee2024compact3dgs}, LightGaussian~\cite{fan2023lightgaussian}, EAGLES~\cite{girish2024eagles}, RadSplat~\cite{niemeyer2024radsplat}, Mini-Splatting~\cite{fang2024minisplatting}, and LP-3DGS~\cite{kim2024lp3dgs}).
In Table~\ref{tab:mipnerf360_comparison}, we additionally report a retrained baseline ($\text{3DGS}^{*}$) to provide a fair comparative reference trained under our identical hardware and schedule.
Note that for LP-3DGS, we evaluate its official implementation but apply max-pooling aggregation instead of its original summation-based method. This provides a stronger baseline and ensures a fair comparison by strictly isolating the effect of the mask activation function.

\paragraph{Evaluation metrics.}
Rendering quality is assessed via PSNR, SSIM~\cite{wang2004image}, and LPIPS~\cite{zhang2018unreasonable}. Compression efficiency is measured by the surviving Gaussian count and storage size (MB).

\paragraph{Implementation details.}
We implement LinearMask-GS in PyTorch, building upon the official 3DGS
codebase~\cite{kerbl20233d}.
The slope $\tau{=}0.1$ (grid search over $\{0.01, 0.05, 0.1, 0.5\}$), minimum-survival ratio $\rho{=}0.30$, and mask weight $\lambda_m{=}5\times10^{-4}$ are tuned on a held-out set and fixed across all scenes; sensitivity analyses are provided in the supplementary material.
Training follows the standard $30{,}000$-iteration 3DGS schedule. We train the 3DGS- and 2DGS-backbone runs on a single NVIDIA RTX~3090 GPU, and the FastGS-backbone run (LinearMask-GS$^{\bigstar}$) on a single NVIDIA RTX~4090 GPU, following the hardware setup of the FastGS paper~\cite{ren2025fastgs}. For a fair wall-clock comparison, the training-time measurements in Table~\ref{tab:mipnerf360_training_time} are all collected on a single RTX~4090.

\subsection{Main Results}
\label{sec:comparison}

We evaluate LinearMask-GS against scene representation and pruning
baselines on three standard benchmarks: Mip-NeRF~360~\cite{barron2022mipnerf360},
Tanks \& Temples~\cite{knapitsch2017tanks}, and Deep
Blending~\cite{hedman2018deepblending}.
For all datasets, we follow the standard evaluation protocol of 3DGS~\cite{kerbl20233d}, where every eighth image is reserved for testing.
We report PSNR (dB), SSIM, LPIPS, and the number of Gaussians (M) on all benchmarks, plus FPS and storage (MB) on Mip-NeRF~360 (measured on the same hardware for reproduced methods, and taken from the original papers for literature baselines). 

\paragraph{Mip-NeRF 360.}
Table~\ref{tab:mipnerf360_comparison} summarizes results across all nine scenes against three families of baselines: the original 3DGS backbone with its retrained reference and the LP-3DGS learned-mask parent; compression- and pruning-oriented systems; and recent fast-training pipelines.
Compared to the vanilla 3DGS~\cite{kerbl20233d} backbone, our method reduces the Gaussian count from 3.36M to 0.94M (a $3.6\times$ reduction) while improving PSNR from 27.21 to 27.70~dB.
Against our primary pruning baseline LP-3DGS~\cite{kim2024lp3dgs}, LinearMask-GS surpasses it by 0.23~dB in PSNR and 0.0135 in SSIM with $1.6\times$ fewer Gaussians, while maintaining comparable rendering speed (591 vs.\ 588~FPS).

Compared to recent compression methods (HAC~\cite{chen2024hac}, HAC++~\cite{chen2025hacpp}) and aggressive pruning baselines (Mini-Splatting~\cite{fang2024minisplatting}, Speedy-Splat~\cite{hanson2025speedy}), LinearMask-GS achieves competitive PSNR with substantially fewer Gaussians than compression methods while maintaining higher quality than the most aggressive pruning baselines (see Table~\ref{tab:mipnerf360_comparison} for details).
For a budget-matched comparison with the most aggressive pruning baseline, we
additionally constrain LinearMask-GS$^{\bigstar}$ to Speedy-Splat's budget of
0.30M Gaussians: over three independent seeds it reaches
$27.40$~dB\,/\,$0.798$\,/\,$0.230$, versus
$26.91$\,/\,$0.781$\,/\,$0.295$ for Speedy-Splat, i.e., better on all three
metrics at the same budget, and still ahead at 0.25M
($27.06$\,/\,$0.791$\,/\,$0.241$); see Sec.~11 of the supplementary material.
Applying LinearMask-GS on the FastGS backbone (LinearMask-GS$^{\bigstar}$, last row), where LinearMask-GS substitutes FastGS's post-densification VCP module, further compresses the FastGS output from 0.40M to 0.38M Gaussians while improving PSNR by 0.13~dB over FastGS (27.69 vs.\ 27.56~dB) and reducing storage to 94.89~MB. This indicates that our masking strategy can replace specialized post-densification pruning components in fast-training pipelines while improving rendering quality.

Figure~\ref{fig:3way_efficiency} summarizes this advantage through efficiency metrics across all three quality dimensions ($10^9 / (\text{LPIPS} \times \#\text{G})$, $\text{PSNR} / \#\text{G}$, $\text{SSIM} / \#\text{G}$), on which LinearMask-GS achieves the highest score against the 3DGS and LP-3DGS reference rows; a per-scene LPIPS-vs-\#G visualization is provided in Sec.~5 of the supplementary material.

\begin{table}[!ht]
\centering
\caption{Quantitative results on Mip-NeRF~360. All 3DGS- and 2DGS-backbone results are obtained under the same RTX~3090 environment as the baselines for a like-for-like comparison; the FastGS-backbone variant ($\bigstar$) follows the FastGS hardware setup (RTX~4090). Baselines: (i) 3DGS and its retrained reference $\text{3DGS}^{*}$, (ii) LP-3DGS evaluated with max-pooling for a matched comparison (Sec.~\ref{subsec:ablation}), and (iii) recent compression, pruning, and fast-training methods. $\bigstar$: our method as a VCP substitute on FastGS. ``--'': metric not reported. Best per column in \textbf{bold}.}
\label{tab:mipnerf360_comparison}
\resizebox{\linewidth}{!}{
\begin{tabular}{lcccccc}
\toprule
Method & LPIPS$\downarrow$ & PSNR$\uparrow$ & SSIM$\uparrow$
       & \#G (M)$\downarrow$ & FPS$\uparrow$ & Storage (MB)$\downarrow$ \\
\midrule
Plenoxels~\cite{fridovich2022plenoxels}
  & 0.463 & 23.08 & 0.626 & --   & 6.79   & 2100.0 \\
INGP-base~\cite{muller2022instant}
  & 0.371 & 25.30 & 0.671 & --   & 11.70  & 13.0   \\
INGP-big~\cite{muller2022instant}
  & 0.331 & 25.59 & 0.699 & --   & 9.43   & 48.0   \\
Mip-NeRF~360~\cite{barron2022mipnerf360}
  & 0.237 & 27.69 & 0.792 & --   & 0.06   & 8.6    \\
\midrule
3DGS~\cite{kerbl20233d}
  & 0.214 & 27.21 & 0.815 & 3.36 & 134.00 & 734.0  \\
3DGS$^{*}$
  & 0.222 & 27.46 & 0.812 & 3.35 & 120.00 & 746.0  \\
LP-3DGS~\cite{kim2024lp3dgs}
  & 0.2258 & 27.47 & 0.8119 & 1.48 & 588.00 & 314.12 \\
\midrule
\multicolumn{7}{l}{\textit{Compression-oriented}} \\
Compact3DGS~\cite{lee2024compact3dgs}
  & 0.247 & 27.08 & 0.798 & 1.39 & 128.00 & 48.8 \\
EAGLES~\cite{girish2024eagles}
  & 0.238 & 27.16 & 0.809 & 1.71 & 131.00 & 54.0   \\
HAC (high-rate)~\cite{chen2024hac}
  & 0.2296 & 27.77 & 0.8109 & 2.26 & --   & 22.94 \\
HAC++ (high-rate)~\cite{chen2025hacpp}
  & 0.2307 & \textbf{27.82} & 0.8109 & 1.85 & --   & \textbf{19.38} \\
\midrule
\multicolumn{7}{l}{\textit{Pruning-oriented}} \\
LightGaussian~\cite{fan2023lightgaussian}
  & 0.231 & 26.54 & 0.798 & 1.05 & 189.00 & 154.5  \\
RadSplat~\cite{niemeyer2024radsplat}
  & 0.218 & 27.08 & 0.811 & 1.92 & 165.00 & 189.2  \\
Mini-Splatting~\cite{fang2024minisplatting}
  & 0.217 & 27.34 & 0.822 & 0.49 & 178.00 & 132.8 \\
PUP 3D-GS~\cite{hanson2025pup}
  & 0.2719 & 26.67 & 0.7862 & --   & 204.81 & 74.65 \\
Speedy-Splat~\cite{hanson2025speedy}
  & 0.295  & 26.91 & 0.781  & \textbf{0.30} & 552.00 & --   \\
\midrule
\multicolumn{7}{l}{\textit{Fast-training systems}} \\
Taming~3DGS~\cite{papantonakis2024taming}
  & 0.261 & 27.48 & 0.794 & 0.68 & 221.00 & --   \\
DashGaussian~\cite{chen2025dashgaussian}
  & 0.218 & 27.73 & 0.817 & 2.40 & 155.00 & --   \\
FastGS~\cite{ren2025fastgs}
  & 0.261 & 27.56 & 0.797 & 0.40 & 579.00 & 99.44 \\
\midrule
\textbf{LinearMask-GS (Ours)}
  & \textbf{0.2095} & 27.70 & \textbf{0.8254}
  & 0.94 & \textbf{591.14} & 226.76 \\
\textbf{LinearMask-GS$^{\bigstar}$ (FastGS backbone)}
  & 0.2227 & 27.69 & 0.8029
  & 0.38 & 344.60 & 94.89 \\
\bottomrule
\end{tabular}}
\end{table}


\paragraph{Tanks \& Temples and Deep Blending.}
On the Tanks~\&~Temples and Deep~Blending benchmarks, LinearMask-GS$^{\bigstar}$ (FastGS-backbone variant that substitutes FastGS's post-densification VCP module with our learned mask) reaches 24.18~dB PSNR and 0.825 SSIM with 0.29M Gaussians on Tanks~\&~Temples, and 30.11~dB PSNR and 0.899 SSIM with 0.29M Gaussians on Deep~Blending. This places LinearMask-GS$^{\bigstar}$ in the sub-million primitive regime of recent fast-training baselines while achieving the best PSNR on Tanks~\&~Temples average (24.18~dB), Dr~Johnson (29.67~dB), and Deep~Blending average (30.11~dB), surpassing FastGS and other fast-training methods on these metrics.
The supplementary material provides the full quantitative comparison against recent fast-training and compact 3DGS baselines, together with per-scene score tables for both benchmarks.

\begin{figure}[!ht]
 \centering
 \includegraphics[width=\linewidth]{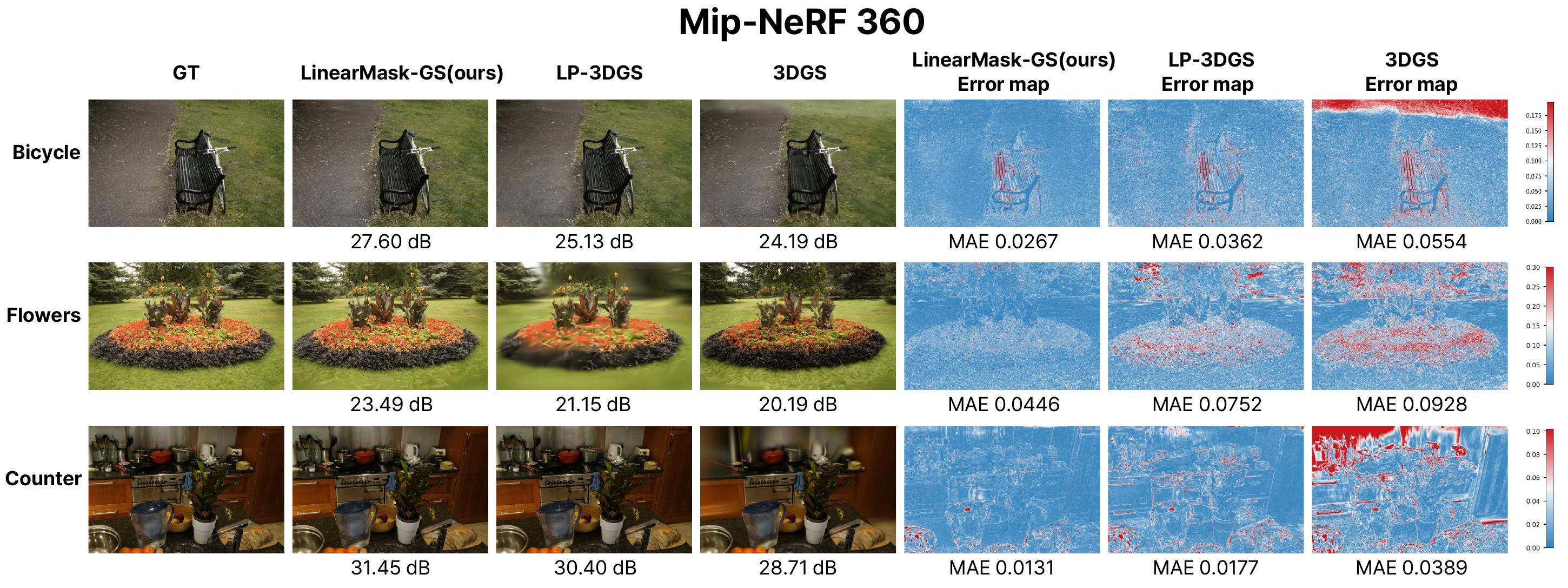}
 \caption{%
  \textbf{Per-scene qualitative comparison} on three representative Mip-NeRF~360 scenes (\textit{Bicycle}, \textit{Flowers}, \textit{Counter}). Each row: GT, renders from LinearMask-GS / LP-3DGS / 3DGS (PSNR below), then per-pixel absolute error maps for the same three (MAE below; scene-specific blue-to-red colormap). LinearMask-GS consistently shows lower error and fewer saturated red regions; remaining six scenes in the supplement.}
 \label{fig:errormap}
\end{figure}

\begin{figure}[!ht]
  \centering
  \includegraphics[width=\linewidth]{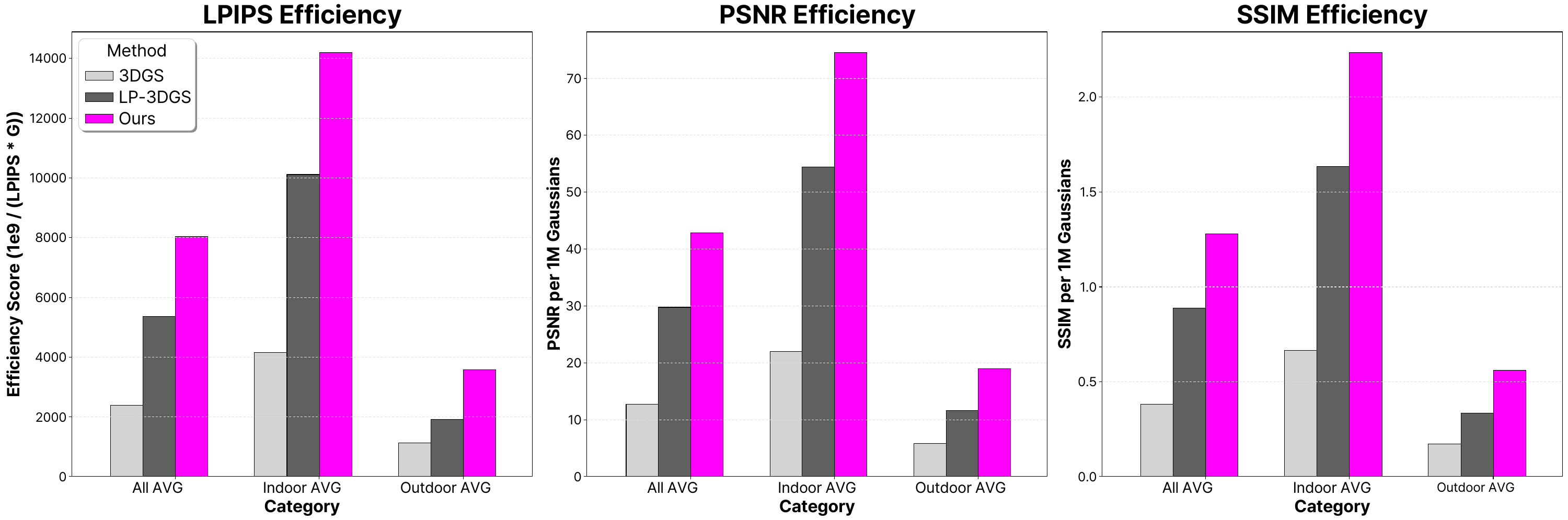}
  \caption{%
      \textbf{Efficiency scores} on Mip-NeRF~360 across three metrics ($10^9 / (\text{LPIPS} \times \#\text{G})$, $\text{PSNR} / \#\text{G}$, $\text{SSIM} / \#\text{G}$). LinearMask-GS achieves the highest score on all three, indicating superior perceptual quality per Gaussian.
    }
  \label{fig:3way_efficiency}
\end{figure}

\paragraph{Qualitative comparison.}
Figure~\ref{fig:errormap} shows per-pixel absolute error maps for three representative Mip-NeRF~360 scenes (\textit{Bicycle}, \textit{Flowers}, \textit{Counter}); error maps for the remaining six scenes are provided in the supplementary material.
For each scene, error maps are visualized under a scene-specific colormap (blue to red indicates low to high absolute error), and per-image PSNR and MAE are reported below each panel.
Across all nine scenes, LinearMask-GS produces lower MAE and fewer saturated high-error (red) regions than both LP-3DGS and 3DGS.
The improvement is most visible in texture-dense outdoor scenes such as \textit{Flowers} and \textit{Stump}, where 3DGS error maps show large red areas around the foliage and 3DGS PSNR drops to 20.19 and 24.94 dB respectively, while LinearMask-GS retains 23.49 and 27.07 dB with substantially cleaner error maps.
In indoor scenes such as \textit{Counter} and \textit{Kitchen}, the per-image PSNR margin over 3DGS exceeds 2.7 dB, reflecting more accurate reconstruction of fine structures like the countertop and the toy loader despite using fewer Gaussians overall (Table~\ref{tab:mipnerf360_comparison}).

\paragraph{Training time.}
Table~\ref{tab:mipnerf360_training_time} reports mean training time over the nine Mip-NeRF~360 scenes, all measured on a single RTX~4090.
On the 3DGS backbone, LinearMask-GS is 2.12 minutes faster than its learned-mask parent LP-3DGS while reaching a much more compact final model with higher PSNR and SSIM (0.94M vs.\ 1.48M Gaussians, Table~\ref{tab:mipnerf360_comparison}), and incurs only a 0.9-minute overhead over the unpruned 3DGS baseline despite the added mask-training and fine-tuning stages.
On the FastGS backbone, LinearMask-GS$^{\bigstar}$ reduces training time (1.93 to 1.81 minutes) while removing additional primitives: the brief mask-learning stage is more than offset by faster iterations on the already-pruned model, yielding a net saving; a full timing breakdown is provided in Sec.~13 of the supplementary material.

\begin{table}[!ht]
\centering
\caption{Training-time comparison on Mip-NeRF~360 (minutes, averaged over nine scenes). All rows are measured on a single RTX~4090. $\bigstar$: our method on FastGS.}
\label{tab:mipnerf360_training_time}
\resizebox{0.7\linewidth}{!}{
\begin{tabular}{lc}
\toprule
Method & Train Time (min)$\downarrow$ \\
\midrule
3DGS~\cite{kerbl20233d} & 12.10 \\
LP-3DGS~\cite{kim2024lp3dgs} & 15.12 \\
\textbf{LinearMask-GS (Ours)} & 13.00 \\
Taming~3DGS~\cite{papantonakis2024taming} & 5.36 \\
DashGaussian~\cite{chen2025dashgaussian} & 6.35 \\
FastGS~\cite{ren2025fastgs} & 1.93 \\
\textbf{LinearMask-GS$^{\bigstar}$ (FastGS backbone)} & \textbf{1.81} \\
\bottomrule
\end{tabular}}
\end{table}

\subsection{Generality across Backbones}
\label{subsec:2dgs}

\paragraph{2DGS.}
We apply LinearMask-GS to the official 2DGS~\cite{huang20242d} codebase to test the generality of our framework across different Gaussian representations. While 3DGS and 2DGS utilize different geometric primitives, they share a consistent alpha-compositing pipeline, allowing our max-pooling score and linear activation to be applied without structural changes. 

As shown in Table~\ref{tab:multibackbone_transfer}, on the nine Mip-NeRF~360 scenes LinearMask-GS reduces the 2DGS Gaussian count by $2.9\times$ (from 1.06M to 0.36M) while improving PSNR by 0.22~dB and SSIM by 0.009, with LPIPS rising only marginally (by 0.008). This indicates that our importance masking successfully identifies redundant 2D disks without sacrificing perceptual quality.


\paragraph{DropGaussian and Octree-GS.}
Beyond 2DGS, we further test LinearMask-GS on two additional Gaussian-based backbones with different representational structures: DropGaussian~\cite{park2025dropgaussian}, which regularizes 3D Gaussians through stochastic dropout, and Octree-GS~\cite{ren2024octreegs}, which uses hierarchical anchor structures.
On DropGaussian, we evaluate five LLFF scenes (9-view) with the masked model repeated over three independent training runs. Averaged over the runs, LinearMask-GS improves PSNR on every scene, by $+0.31$ to $+0.52$~dB over the unpruned backbone, while retaining $41\%$ fewer primitives than the Gumbel variant (per-scene results in Sec.~12 of the supplementary material). The single \textit{fern} run of Table~\ref{tab:multibackbone_transfer} gives a smaller gain ($+0.03$~dB at a $2.8\times$ reduction), indicating that single runs of this sparse-view backbone are noisy.
On Octree-GS, the \textit{amsterdam} scene from Bungee-NeRF shows a $2.9\times$ anchor reduction (1.09M to 0.38M) with $+0.12$~dB PSNR and SSIM/LPIPS within $0.002$ of the reference; on the \textit{bicycle} and \textit{garden} scenes of Mip-NeRF~360, anchors drop by $40.2\%$ relative to the Gumbel variant at a mean PSNR change of $-0.11$~dB. Across all three Octree-GS scenes, quality thus moves by at most $\approx$0.1~dB in either direction while anchors are reduced by $40\%$ or more.
Table~\ref{tab:multibackbone_transfer} consolidates these results alongside the 2DGS comparison, and additionally reports the 2DGS variant with the linear-increment activation replaced by Gumbel-Sigmoid; the activation-level comparison across backbones is analyzed in Sec.~\ref{subsec:ablation}.

\begin{table}[!ht]
\centering
\caption{Generality of LinearMask-GS across alternative Gaussian-based backbones, applied as a downstream pruning stage. \#Primitives counts Gaussians for 2DGS and DropGaussian, and anchors for Octree-GS. 2DGS results average over the nine Mip-NeRF~360 scenes; DropGaussian uses \textit{fern} (LLFF) and Octree-GS uses \textit{amsterdam} (Bungee-NeRF). ``+ LP-3DGS (Gumbel)'' matches ``+ LinearMask-GS'' except in activation (see Sec.~\ref{subsec:2dgs}). Best per column in \textbf{bold}.}
\label{tab:multibackbone_transfer}
\resizebox{\linewidth}{!}{%
\begin{tabular}{llrrrr}
\toprule
Backbone & Method & LPIPS$\downarrow$ & PSNR$\uparrow$ & SSIM$\uparrow$ & \#Primitives$\downarrow$ \\
\midrule
\multirow{3}{*}{2DGS~\cite{huang20242d}}
  & 2DGS (no pruning)               & \textbf{0.2586} & 26.7390 & 0.7935 & 1{,}063{,}924 \\
  & + LP-3DGS (Gumbel)              & 0.2734 & 26.9261 & 0.8022 & 628{,}840 \\
  & \textbf{+ LinearMask-GS$^{*}$}  & 0.2663 & \textbf{26.9583} & \textbf{0.8025} & \textbf{364{,}448} \\
\midrule
\multirow{3}{*}{DropGaussian~\cite{park2025dropgaussian}}
  & DropGaussian (no pruning)       & 0.1106 & 26.6495 & 0.8806 & 104{,}877 \\
  & + LP-3DGS (Gumbel)              & 0.1098 & 26.6658 & 0.8813 & 62{,}182 \\
  & \textbf{+ LinearMask-GS}        & \textbf{0.1087} & \textbf{26.6756} & \textbf{0.8815} & \textbf{37{,}433} \\
\midrule
\multirow{3}{*}{Octree-GS~\cite{ren2024octreegs}}
  & Octree-GS (no pruning)          & \textbf{0.0917} & 27.8570 & 0.9184 & 1{,}094{,}939 \\
  & + LP-3DGS (Gumbel)              & 0.0908 & 27.9280 & \textbf{0.9192} & 648{,}221 \\
  & \textbf{+ LinearMask-GS}        & 0.0928 & \textbf{27.9720} & 0.9191 & \textbf{380{,}726} \\
\bottomrule
\end{tabular}}
\end{table}

\subsection{Ablation Study}
\label{subsec:ablation}

We ablate the core design choices of LinearMask-GS on the Mip-NeRF~360 dataset:
the importance aggregation, the mask activation, and the minimum-survival ratio $\rho$.
Unless stated otherwise, all variants use the default configuration:
$\lambda_m = 5\times10^{-4}$, $\tau = 0.1$, and $\rho = 0.30$.
Detailed analyses of training convergence and additional hyperparameter sensitivity (sparsity weight $\lambda_m$ and slope $\tau$) are provided in the supplementary material.

\paragraph{Aggregation and activation.}
Table~\ref{tab:ablation_2x2} presents a full factorial comparison
of summation-based~\cite{fang2024minisplatting} and
max-pooling~\cite{niemeyer2024radsplat} aggregation against
Gumbel-Sigmoid~\cite{kim2024lp3dgs} and linear increment activations.
Note that the ``Max + Gumbel-Sigmoid'' variant corresponds exactly to the LP-3DGS baseline reported in Table~\ref{tab:mipnerf360_comparison}. 
For the summation-based variants (rows~1--2), we use $\rho = 0.40$
following the pruning statistics of LP-3DGS~\cite{kim2024lp3dgs};
max-pooling variants use the default $\rho = 0.30$.
Under max-pooling, replacing Gumbel-Sigmoid with the linear increment activation reduces the Gaussian count from 1.48M to 0.94M and improves both PSNR (27.47 to 27.70 dB) and LPIPS (0.2258 to 0.2095). Under summation, the linear-increment variant reduces the count from 1.43M to 1.26M with nearly identical PSNR/LPIPS, indicating that the view-count inflation introduced by summation aggregation dominates the effect of the activation choice and obscures the stability benefit that Linear provides. The combination of max-pooling aggregation and the linear-increment activation gives the best operating point and is used in LinearMask-GS.

\paragraph{Ranking reliability.}
The premise of top-$\rho N$ pruning is that the ranking induced by $\psi_i$
tracks importance, so we test this directly: with hard pruning disabled, the
mask ranking through the window is compared against a held-out proxy,
the max-pooled contribution of each Gaussian (Eq.~\eqref{eq:max_score}) at
iteration 30k (protocol in Sec.~3 of the supplementary material). Across all
nine Mip-NeRF~360 scenes, at the pruning decision point:
(i)~re-sampling the Gumbel noise alone flips $47.3\%$ of keep/prune
decisions ($46$--$48\%$ on every scene; rank correlation between two noise
draws ${<}\,0.1$), so the sampled decision is governed by activation noise
rather than by a stable learned ranking, whereas the linear mask is
deterministic ($0\%$ flips);
(ii)~$52.2\%$ of Gumbel-Sigmoid mask values are saturated under this
protocol versus $0.00\%$ for linear, consistent with the $51.19\%$ observed
under the standard schedule (Sec.~\ref{subsec:linear_mask});
(iii)~the Spearman correlation between the mask ranking and the proxy is
$0.29$ for linear versus $0.16$ for Gumbel-Sigmoid ($1.8\times$ higher
overall and $2.8\times$ on outdoor scenes; per-iteration curves in the
supplementary material).

\paragraph{Contribution of the learned mask.}
To isolate what the mask adds over its ingredients, we also prune directly by the raw score: same schedule, top-$30\%$ by $S_i$ at 20k, then identical fine-tuning (``Max + none'' in Table~\ref{tab:ablation_2x2}). This mask-free baseline is strong ($27.57$~dB\,/\,$0.8231$\,/\,$0.2143$ at $0.96$M), yet our learned mask is ahead on eight of the nine scenes at a smaller budget ($+0.13$~dB at $0.94$M; the only exception is \textit{Treehill}; per-scene results in Sec.~10 of the supplement).
The Gumbel-Sigmoid mask falls \emph{below} it ($27.47$~dB, $1.48$M): once the ranking saturates, a learned mask is worse than none. With the ranking-reliability analysis above, this locates the gain: not max-pooling or fine-tuning, both shared with these baselines, but the stability of the ranking itself.

\paragraph{Minimum-survival ratio $\rho$.}
We sweep $\rho \in \{0.10,\,0.20,\,0.30,\,0.40,\,0.50\}$ on all nine
Mip-NeRF~360 scenes for both the linear-increment activation and the matched
Gumbel-Sigmoid baseline under the identical top-$\rho N$ rule of
Eq.~\eqref{eq:hard_pruning} (Secs.~8--9 of the supplementary material). Two
observations follow. First, the two quality--compression curves share the
same \emph{shape}: quality improves steeply up to $\rho{=}0.30$ and largely
saturates beyond it (over $\rho{=}0.30{\to}0.50$, PSNR gains total
$+0.04$~dB for linear and $+0.11$~dB for Gumbel-Sigmoid). Within the linear
sweep, $\rho{=}0.20$ removes too many meaningful primitives ($-0.28$~dB),
whereas $\rho{=}0.40$ retains $34\%$ more Gaussians for only $+0.03$~dB, so
we keep $\rho{=}0.30$ as the default. Second, the curves differ in
\emph{position}: because top-$\rho N$ retains the same fraction of a
near-identical primitive pool, equal $\rho$ implies matched budgets (\#G
differs by $\leq 0.01$M), and at every $\rho$ the linear-increment
activation sits $+0.40$--$0.48$~dB above the Gumbel-Sigmoid baseline at the
same budget. The gap does not close anywhere on the sweep: the baseline's
best point ($27.33$~dB at $\rho{=}0.50$, $1.57$M) falls below even the
linear $\rho{=}0.20$ setting ($27.42$~dB at $0.63$M, $2.5\times$ fewer
primitives) and never reaches our default ($27.70$~dB at $0.94$M) despite
$67\%$ more primitives. The stable ranking thus does not reshape the
trade-off; it selects a better subset at every budget, shifting the entire
curve. 

\begin{table}[!ht]
\centering
\caption{
  Full factorial ablation of importance aggregation and mask
  activation on Mip-NeRF~360, extended with a mask-free baseline that
  prunes directly by the raw max-pooled score $S_i$ (``none'').
  \checkmark~marks the components used in LinearMask-GS.
  Best results are in \textbf{bold}.
}
\label{tab:ablation_2x2}
{\small
\renewcommand{\arraystretch}{0.92}
\begin{tabular}{ll cccc}
\toprule
Aggregation & Activation
  & PSNR$\uparrow$ & SSIM$\uparrow$
  & LPIPS$\downarrow$ & \#G (M)$\downarrow$ \\
\midrule
Sum~\cite{fang2024minisplatting}
  & Gumbel-Sigmoid~\cite{kim2024lp3dgs}
  & 27.12 & 0.805  & 0.239  & 1.43 \\
Sum~\cite{fang2024minisplatting}
  & Linear increment
  & 27.05 & 0.806  & 0.238  & 1.26 \\
Max~\cite{niemeyer2024radsplat}
  & None (top-$\rho N$ by $S_i$)
  & 27.57 & 0.8231 & 0.2143 & 0.96 \\
Max~\cite{niemeyer2024radsplat}
  & Gumbel-Sigmoid~\cite{kim2024lp3dgs}
  & 27.47 & 0.8119 & 0.2258 & 1.48 \\
Max\,\checkmark~\cite{niemeyer2024radsplat}
  & Linear increment\,\checkmark
  & \textbf{27.70} & \textbf{0.8254}
  & \textbf{0.2095} & \textbf{0.94} \\
\bottomrule
\end{tabular}}
\end{table}

\paragraph{Activation choice across backbones.}
The factorial ablation above is conducted on the 3DGS backbone. To verify that the activation itself, rather than the masking framework, drives the gain, we compare the two activations on DropGaussian and Octree-GS with the mask window, sparsity loss, importance score, and target keep ratio held identical (the ``+ LP-3DGS (Gumbel)'' vs.\ ``+ LinearMask-GS'' rows of Table~\ref{tab:multibackbone_transfer}). The linear-increment activation prunes $40\%$ (DropGaussian, \textit{fern}) and $41\%$ (Octree-GS, \textit{amsterdam}) further than the Gumbel-Sigmoid variant while maintaining or improving quality; since these backbones use markedly different mask-training schedules (1{,}500 and 4{,}000 iterations), the activation choice, not merely the masking framework, governs the achievable quality--compression trade-off.

\section{Conclusion}
\label{sec:conclusion}

In this work, we identified a limitation of existing learned-mask pruning for 3D Gaussian Splatting and introduced LinearMask-GS to address it. The steep slope of Gumbel-Sigmoid pushes mask values to $0$ or $1$ before the importance ranking stabilizes, a failure that longer mask-training windows exacerbate rather than fix, producing a two-peaked distribution from which that ranking can no longer be reliably recovered. Replacing it with a linear-increment activation keeps mask values in the mid-confidence regime and preserves their ranking, so a simple top-$\rho N$ selection suffices, and this single change transfers without modification across diverse Gaussian-based backbones. 

\paragraph{Limitations.}
The slope $\tau$ and ratio $\rho$ are tuned on a held-out set (Sec.~\ref{subsec:setup}) and fixed across scenes; the brief mask window can leave specular or transparent regions under-discriminated, and the global $\rho$ sets an overall rather than per-region budget. The method also inherits the static-scene assumption of its backbones; scene-adaptive scheduling and dynamic-scene extensions are left for future work. 

\section*{Acknowledgments}
This work was partly supported by the National Research Foundation of Korea (NRF) grant funded by the Korea government (MSIT) (RS-2024-00337250), by the Korea Institute for Advancement of Technology (KIAT) grant funded by the Korea government (MOTIE) (P0023718, Inorganic Light-emitting Display Expert Training Program for Display Technology Transition), and by the Institute of Information \& Communications Technology Planning \& Evaluation (IITP, AI Computing Support Project for R\&D) grant funded by the Korea government (MSIT) (RS-2026-25505492, High-Performance Research AI Computing Infrastructure Support at the 2 PFLOPS Scale).

\bibliography{main}

\begin{thebibliography}{31}
\providecommand{\natexlab}[1]{#1}
\providecommand{\url}[1]{\texttt{#1}}
\expandafter\ifx\csname urlstyle\endcsname\relax
  \providecommand{\doi}[1]{doi: #1}\else
  \providecommand{\doi}{doi: \begingroup \urlstyle{rm}\Url}\fi

\bibitem[Barron et~al.(2022)Barron, Mildenhall, Verbin, Srinivasan, and
  Hedman]{barron2022mipnerf360}
Jonathan~T. Barron, Ben Mildenhall, Dor Verbin, Pratul~P. Srinivasan, and Peter
  Hedman.
\newblock Mip-{NeRF} 360: Unbounded anti-aliased neural radiance fields.
\newblock In \emph{CVPR}, pages 5470--5479, 2022.

\bibitem[Chen et~al.(2024)Chen, Wu, Lin, Harandi, and Cai]{chen2024hac}
Yihang Chen, Qianyi Wu, Weiyao Lin, Mehrtash Harandi, and Jianfei Cai.
\newblock {HAC}: Hash-grid assisted context for {3D} gaussian splatting
  compression.
\newblock \emph{arXiv preprint arXiv:2403.14530}, 2024.

\bibitem[Chen et~al.(2025{\natexlab{a}})Chen, Wu, Lin, Harandi, and
  Cai]{chen2025hacpp}
Yihang Chen, Qianyi Wu, Weiyao Lin, Mehrtash Harandi, and Jianfei Cai.
\newblock {HAC++}: Towards 100x compression of {3D} gaussian splatting.
\newblock \emph{arXiv preprint arXiv:2501.12255}, 2025{\natexlab{a}}.

\bibitem[Chen et~al.(2025{\natexlab{b}})Chen, Jiang, Jiang, Tang, Li, Liu, and
  Nie]{chen2025dashgaussian}
Youyu Chen, Junjun Jiang, Kui Jiang, Xiao Tang, Zhihao Li, Xianming Liu, and
  Yinyu Nie.
\newblock {DashGaussian}: Optimizing {3D} gaussian splatting in 200 seconds.
\newblock \emph{arXiv preprint arXiv:2503.18402}, 2025{\natexlab{b}}.

\bibitem[Fan et~al.(2024)Fan, Wang, Wen, Zhu, Xu, and
  Wang]{fan2023lightgaussian}
Zhiwen Fan, Kevin Wang, Kairun Wen, Zehao Zhu, Dejia Xu, and Zhangyang Wang.
\newblock {LightGaussian}: Unbounded {3D} gaussian compression with 15x
  reduction and 200+ {FPS}.
\newblock In \emph{NeurIPS}, pages 140138--140158, 2024.

\bibitem[Fang and Wang(2024)]{fang2024minisplatting}
Guangchi Fang and Bing Wang.
\newblock Mini-splatting: Representing scenes with a constrained number of
  gaussians.
\newblock In \emph{ECCV}, pages 165--181, 2024.

\bibitem[Fridovich-Keil et~al.(2022)Fridovich-Keil, Yu, Tancik, Chen, Recht,
  and Kanazawa]{fridovich2022plenoxels}
Sara Fridovich-Keil, Alex Yu, Matthew Tancik, Qinhong Chen, Benjamin Recht, and
  Angjoo Kanazawa.
\newblock Plenoxels: Radiance fields without neural networks.
\newblock In \emph{CVPR}, pages 5501--5510, 2022.

\bibitem[Girish et~al.(2024)Girish, Gupta, and Shrivastava]{girish2024eagles}
Sharath Girish, Kamal Gupta, and Abhinav Shrivastava.
\newblock {EAGLES}: Efficient accelerated {3D} gaussians with lightweight
  encodings.
\newblock In \emph{ECCV}, pages 54--71, 2024.

\bibitem[Hanson et~al.(2025{\natexlab{a}})Hanson, Tu, Lin, Singla, Zwicker, and
  Goldstein]{hanson2025speedy}
Alex Hanson, Allen Tu, Geng Lin, Vasu Singla, Matthias Zwicker, and Tom
  Goldstein.
\newblock {Speedy-Splat}: Fast {3D} gaussian splatting with sparse pixels and
  sparse primitives.
\newblock \emph{arXiv preprint arXiv:2412.00578}, 2025{\natexlab{a}}.

\bibitem[Hanson et~al.(2025{\natexlab{b}})Hanson, Tu, Singla, Jayawardhana,
  Zwicker, and Goldstein]{hanson2025pup}
Alex Hanson, Allen Tu, Vasu Singla, Mayuka Jayawardhana, Matthias Zwicker, and
  Tom Goldstein.
\newblock {PUP 3D-GS}: Principled uncertainty pruning for {3D} gaussian
  splatting.
\newblock \emph{arXiv preprint arXiv:2406.10219}, 2025{\natexlab{b}}.

\bibitem[Hedman et~al.(2018)Hedman, Philip, Price, Frahm, Drettakis, and
  Brostow]{hedman2018deepblending}
Peter Hedman, Julien Philip, True Price, Jan-Michael Frahm, George Drettakis,
  and Gabriel Brostow.
\newblock Deep blending for free-viewpoint image-based rendering.
\newblock \emph{ACM TOG}, 37\penalty0 (6):\penalty0 1--15, 2018.

\bibitem[Huang et~al.(2024)Huang, Yu, Chen, Geiger, and Gao]{huang20242d}
Binbin Huang, Zehao Yu, Anpei Chen, Andreas Geiger, and Shenghua Gao.
\newblock {2D} gaussian splatting for geometrically accurate radiance fields.
\newblock In \emph{ACM SIGGRAPH Conference Papers}, pages 1--11, 2024.

\bibitem[Kerbl et~al.(2023)Kerbl, Kopanas, Leimk{\"u}hler, Drettakis,
  et~al.]{kerbl20233d}
Bernhard Kerbl, Georgios Kopanas, Thomas Leimk{\"u}hler, George Drettakis,
  et~al.
\newblock {3D} gaussian splatting for real-time radiance field rendering.
\newblock \emph{ACM TOG}, 42\penalty0 (4):\penalty0 139--1, 2023.

\bibitem[Knapitsch et~al.(2017)Knapitsch, Park, Zhou, and
  Koltun]{knapitsch2017tanks}
Arno Knapitsch, Jaesik Park, Qian-Yi Zhou, and Vladlen Koltun.
\newblock Tanks and temples: Benchmarking large-scale scene reconstruction.
\newblock \emph{ACM TOG}, 36\penalty0 (4):\penalty0 1--13, 2017.

\bibitem[Lee et~al.(2024)Lee, Rho, Sun, Ko, and Park]{lee2024compact3dgs}
Joo~Chan Lee, Daniel Rho, Xiangyu Sun, Jong~Hwan Ko, and Eunbyung Park.
\newblock Compact {3D} gaussian splatting for static and dynamic radiance
  fields.
\newblock \emph{arXiv preprint arXiv:2408.03822}, 2024.

\bibitem[Loshchilov and Hutter(2017)]{loshchilov2018decoupled}
Ilya Loshchilov and Frank Hutter.
\newblock Decoupled weight decay regularization.
\newblock \emph{arXiv preprint arXiv:1711.05101}, 2017.

\bibitem[Lu et~al.(2024)Lu, Yu, Xu, Xiangli, Wang, Lin, and
  Dai]{lu2024scaffold}
Tao Lu, Mulin Yu, Linning Xu, Yuanbo Xiangli, Limin Wang, Dahua Lin, and
  Bo~Dai.
\newblock Scaffold-{GS}: Structured {3D} gaussians for view-adaptive rendering.
\newblock In \emph{CVPR}, pages 20654--20664, 2024.

\bibitem[Mallick et~al.(2024)Mallick, Goel, Kerbl, Steinberger, Carrasco, and
  De~La~Torre]{papantonakis2024taming}
Saswat~Subhajyoti Mallick, Rahul Goel, Bernhard Kerbl, Markus Steinberger,
  Francisco~Vicente Carrasco, and Fernando De~La~Torre.
\newblock Taming {3DGS}: High-quality radiance fields with limited resources.
\newblock In \emph{ACM SIGGRAPH Asia Conference Papers}, pages 1--11, 2024.

\bibitem[Mildenhall et~al.(2021)Mildenhall, Srinivasan, Tancik, Barron,
  Ramamoorthi, and Ng]{mildenhall2021nerf}
Ben Mildenhall, Pratul~P Srinivasan, Matthew Tancik, Jonathan~T Barron, Ravi
  Ramamoorthi, and Ren Ng.
\newblock {NeRF}: Representing scenes as neural radiance fields for view
  synthesis.
\newblock \emph{Communications of the ACM}, 65\penalty0 (1):\penalty0 99--106,
  2021.

\bibitem[M{\"u}ller et~al.(2022)M{\"u}ller, Evans, Schied, and
  Keller]{muller2022instant}
Thomas M{\"u}ller, Alex Evans, Christoph Schied, and Alexander Keller.
\newblock Instant neural graphics primitives with a multiresolution hash
  encoding.
\newblock \emph{ACM TOG}, 41\penalty0 (4):\penalty0 1--15, 2022.

\bibitem[Niedermayr et~al.(2024)Niedermayr, Stumpfegger, and
  Westermann]{niedermayr2024compressed3dgs}
Simon Niedermayr, Josef Stumpfegger, and R{\"u}diger Westermann.
\newblock Compressed {3D} gaussian splatting for accelerated novel view
  synthesis.
\newblock In \emph{CVPR}, pages 10349--10358, 2024.

\bibitem[Niemeyer et~al.(2025)Niemeyer, Manhardt, Rakotosaona, Oechsle,
  Duckworth, Gosula, Tateno, Bates, Kaeser, and Tombari]{niemeyer2024radsplat}
Michael Niemeyer, Fabian Manhardt, Marie-Julie Rakotosaona, Michael Oechsle,
  Daniel Duckworth, Rama Gosula, Keisuke Tateno, John Bates, Dominik Kaeser,
  and Federico Tombari.
\newblock {RadSplat}: Radiance field-informed gaussian splatting for robust
  real-time rendering with 900+ {FPS}.
\newblock In \emph{International Conference on 3D Vision (3DV)}, pages
  134--144, 2025.

\bibitem[Papantonakis et~al.(2024)Papantonakis, Kopanas, Kerbl, Lanvin, and
  Drettakis]{papantonakis2024reduced}
Panagiotis Papantonakis, Georgios Kopanas, Bernhard Kerbl, Alexandre Lanvin,
  and George Drettakis.
\newblock Reducing the memory footprint of {3D} gaussian splatting.
\newblock \emph{Proceedings of the ACM on Computer Graphics and Interactive
  Techniques}, 7\penalty0 (1):\penalty0 1--17, 2024.

\bibitem[Park et~al.(2025)Park, Ryu, and Kim]{park2025dropgaussian}
Hyunwoo Park, Gun Ryu, and Wonjun Kim.
\newblock {DropGaussian}: Structural regularization for sparse-view gaussian
  splatting.
\newblock In \emph{Proceedings of the IEEE/CVF Conference on Computer Vision
  and Pattern Recognition (CVPR)}, pages 21600--21609, 2025.

\bibitem[Ren et~al.(2024)Ren, Jiang, Lu, Yu, Xu, Ni, and Dai]{ren2024octreegs}
Kerui Ren, Lihan Jiang, Tao Lu, Mulin Yu, Linning Xu, Zhangkai Ni, and Bo~Dai.
\newblock {Octree-GS}: Towards consistent real-time rendering with
  {LOD}-structured {3D} gaussians.
\newblock \emph{arXiv preprint arXiv:2403.17898}, 2024.

\bibitem[Ren et~al.(2025)Ren, Wen, Fang, and Lu]{ren2025fastgs}
Shiwei Ren, Tianci Wen, Yongchun Fang, and Biao Lu.
\newblock {FastGS}: Training {3D} gaussian splatting in 100 seconds.
\newblock \emph{arXiv preprint arXiv:2511.04283}, 2025.

\bibitem[Wang et~al.(2004)Wang, Bovik, Sheikh, and Simoncelli]{wang2004image}
Zhou Wang, Alan~C Bovik, Hamid~R Sheikh, and Eero~P Simoncelli.
\newblock Image quality assessment: From error visibility to structural
  similarity.
\newblock \emph{IEEE TIP}, 13\penalty0 (4):\penalty0 600--612, 2004.

\bibitem[Xie et~al.(2024)Xie, Zhang, Tang, Bai, Lu, Ge, and
  Wang]{xie2024mesongs}
Shuzhao Xie, Weixiang Zhang, Chen Tang, Yunpeng Bai, Rongwei Lu, Shijia Ge, and
  Zhi Wang.
\newblock {MesonGS}: Post-training compression of {3D} gaussians via efficient
  attribute transformation.
\newblock In \emph{ECCV}, pages 434--452, 2024.

\bibitem[Yu et~al.(2024)Yu, Chen, Huang, Sattler, and
  Geiger]{yu2024mipsplatting}
Zehao Yu, Anpei Chen, Binbin Huang, Torsten Sattler, and Andreas Geiger.
\newblock Mip-splatting: Alias-free {3D} gaussian splatting.
\newblock In \emph{CVPR}, pages 19447--19456, 2024.

\bibitem[Zhang et~al.(2018)Zhang, Isola, Efros, Shechtman, and
  Wang]{zhang2018unreasonable}
Richard Zhang, Phillip Isola, Alexei~A Efros, Eli Shechtman, and Oliver Wang.
\newblock The unreasonable effectiveness of deep features as a perceptual
  metric.
\newblock In \emph{CVPR}, pages 586--595, 2018.

\bibitem[Zhang et~al.(2024)Zhang, Song, Lee, Yang, Peng, Chellappa, and
  Fan]{kim2024lp3dgs}
Zhaoliang Zhang, Tianchen Song, Yongjae Lee, Li~Yang, Cheng Peng, Rama
  Chellappa, and Deliang Fan.
\newblock {LP-3DGS}: Learning to prune {3D} gaussian splatting.
\newblock In \emph{NeurIPS}, pages 122434--122457, 2024.

\end{thebibliography}

\clearpage


\section*{Supplementary material}

# Supplementary Material

## LinearMask-GS: Stable-Mask Importance Pruning for Compact 3D Gaussian Splatting

Donghun Ryu[1]
donghun0621@cau.ac.kr
Minhyeok Lee[1,2†]
mlee@cau.ac.kr

[1] Department of Intelligent Semiconductor Engineering
Chung-Ang University
Seoul, Republic of Korea

[2] School of Electrical and Electronics Engineering
Chung-Ang University
Seoul, Republic of Korea

This document provides supplementary material referenced in the main paper. It includes additional analyses of training convergence, hyperparameter sensitivity, and mask-entropy evolution; a quantitative analysis of the reliability of the learned mask ranking and of the effect of the mask-window length and placement; per-scene quantitative results on Mip-NeRF 360; the full Tanks & Temples and Deep Blending comparison; per-scene qualitative comparisons and minimum-survival-ratio $\rho$ sweeps for both the linear-increment activation and the matched Gumbel-Sigmoid baseline; a mask-free pruning baseline, a budget-matched comparison with Speedy-Splat, and an extended cross-backbone evaluation; a decomposition of the mask-learning speed gap on the FastGS backbone; and a description of the accompanying video.

# 1 Additional Experimental Results

In this section, we present additional analyses supporting LinearMask-GS, including training convergence profiles and sensitivity analyses for key hyperparameters.

## 1.1 Training convergence

Figure 1 plots PSNR and SSIM from iteration 19,500 onward. LinearMask-GS reaches higher PSNR and SSIM than LP-3DGS within the 500-iteration mask-training phase (iter. 19,500–20,000) and maintains this advantage through fine-tuning. The faster convergence reflects the gentle slope of the linear-increment activation, which keeps mask values away from the saturation boundaries that Gumbel-Sigmoid reaches early in the mask-training window. As discussed in the main paper, this consistent gradient flow translates to substantially better efficiency scores across all metric dimensions.


† Corresponding author.

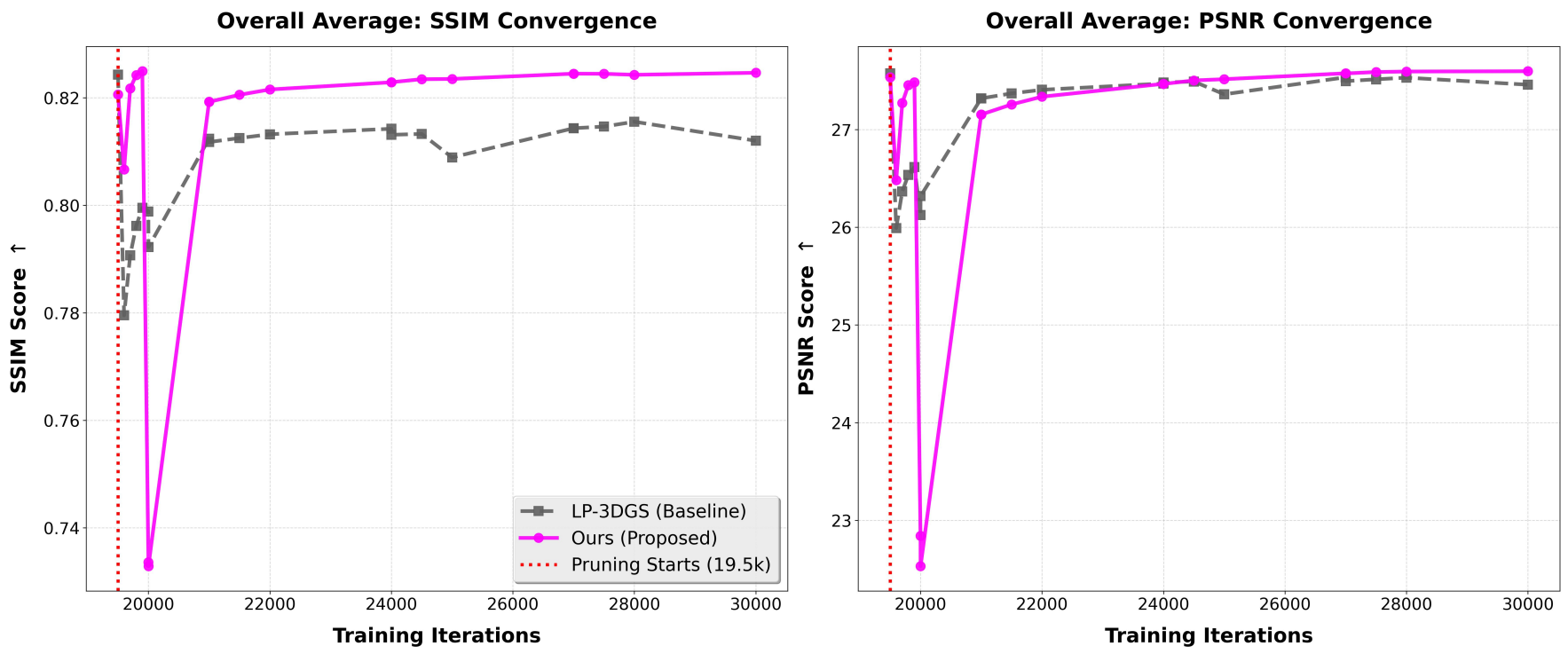


Figure 1: **Training convergence.** PSNR (dB) and SSIM after iteration 19,500 on Mip-NeRF 360. LinearMask-GS converges faster and to higher values than LP-3DGS.

## 1.2 Sensitivity to sparsity weight $\lambda_m$

Figure 2 evaluates $\lambda_m \in \{10^{-4}, 5\times10^{-4}, 10^{-3}, 5\times10^{-3}\}$, using the same per-Gaussian efficiency scores ($10^9$/(LPIPS × #G), PSNR/#G, SSIM/#G) as Figure 3 of the main paper. The default $\lambda_m = 5\times10^{-4}$ achieves the best LPIPS-based efficiency score. Larger values increase sparsity at the cost of perceptual quality, while smaller values provide insufficient regularization; all variants with $\lambda_m \in [10^{-4}, 5\times10^{-4}]$ robustly outperform LP-3DGS.

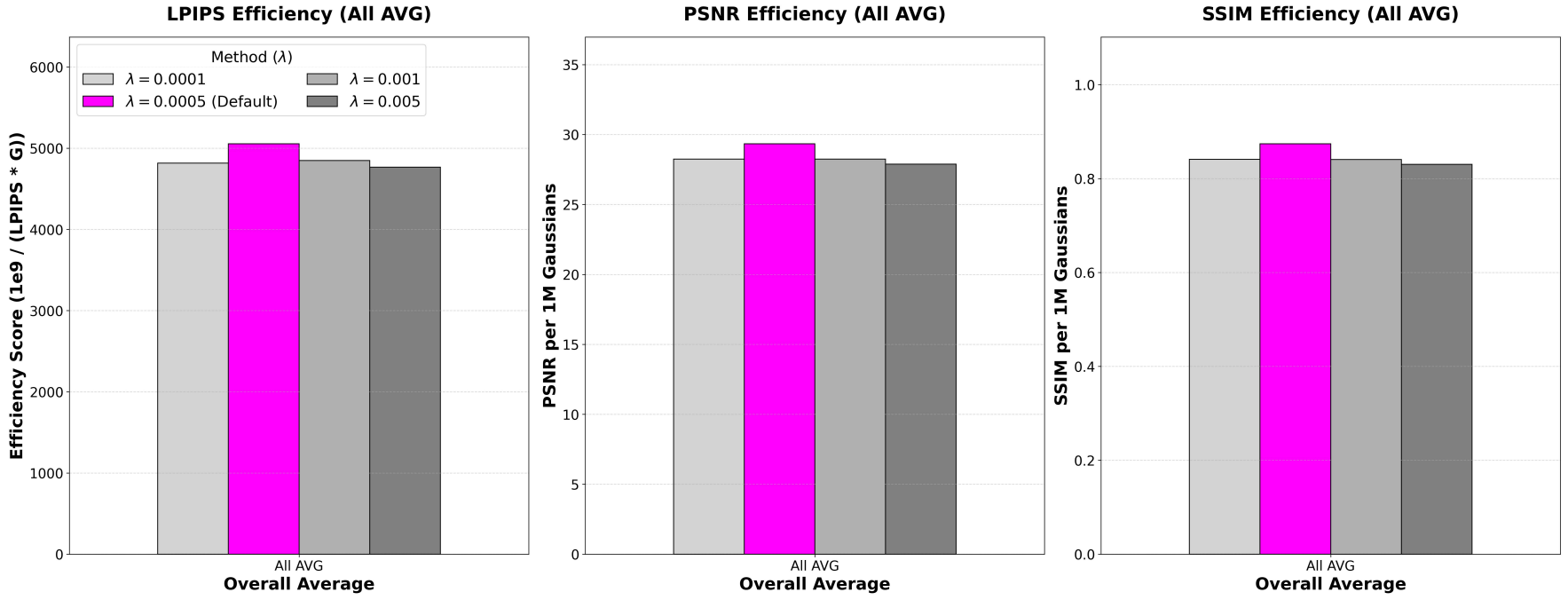


Figure 2: **Sensitivity to $\lambda_m$.** Efficiency score under varying $\lambda_m$ on Mip-NeRF 360. The default $\lambda_m = 5\times10^{-4}$ (highlighted) yields the best LPIPS-based efficiency score.

## 1.3 Sensitivity to slope $\tau$

The main paper reports the primary grid search over $\tau \in \{0.01, 0.05, 0.1, 0.5\}$. Figure 3 evaluates an extended range $\tau \in \{0.01, 0.05, 0.1, 0.25, 0.5, 1.0\}$. The linear-increment activation is robust to the choice of slope: efficiency scores remain high across $0.01 \le \tau \le 0.5$, and in PSNR terms every $\tau \in [0.05, 0.5]$ stays within 0.1 dB of the default. The default $\tau = 0.1$ gives the best balance, while only extreme values (e.g., $\tau = 1.0$) show a noticeable reduction in efficiency.

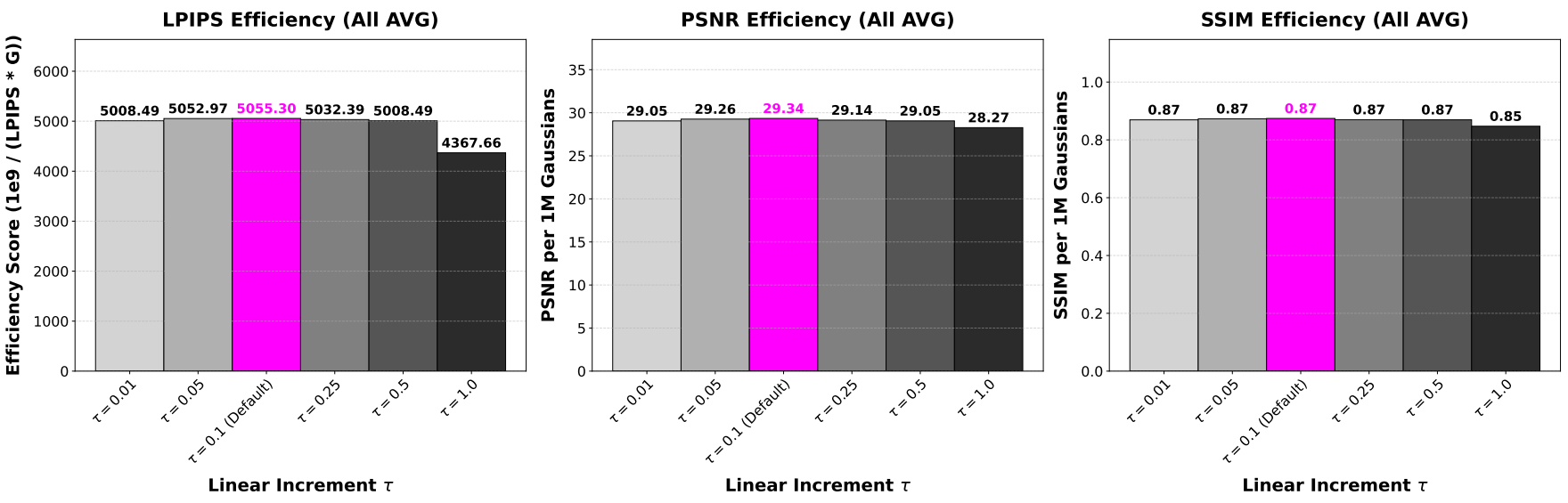


Figure 3: **Sensitivity to $\tau$.** Efficiency score under varying slope $\tau$ on Mip-NeRF 360. The default $\tau = 0.1$ (highlighted) achieves the best LPIPS-based efficiency score.

# 2 Prevention of Premature Binarization

A key advantage of LinearMask-GS is the prevention of premature binarization, where the Gumbel-Sigmoid activation forcefully pushes mask values toward 0 or 1 before the model has adequately evaluated their contribution to the scene. To empirically validate this, we analyze the evolution of mask entropy during the mask-training phase. The binary entropy of a per-Gaussian mask value $\psi_i$ is

$$H(\psi_i) = -\psi_i \log \psi_i - (1-\psi_i)\log(1-\psi_i). \tag{1}$$

A higher entropy indicates that mask values remain concentrated around 0.5, so the corresponding Gaussians are still under active evaluation; a near-zero entropy implies the values have collapsed to the extremes (0 or 1), committing to a prune/keep decision before that evaluation has stabilized.

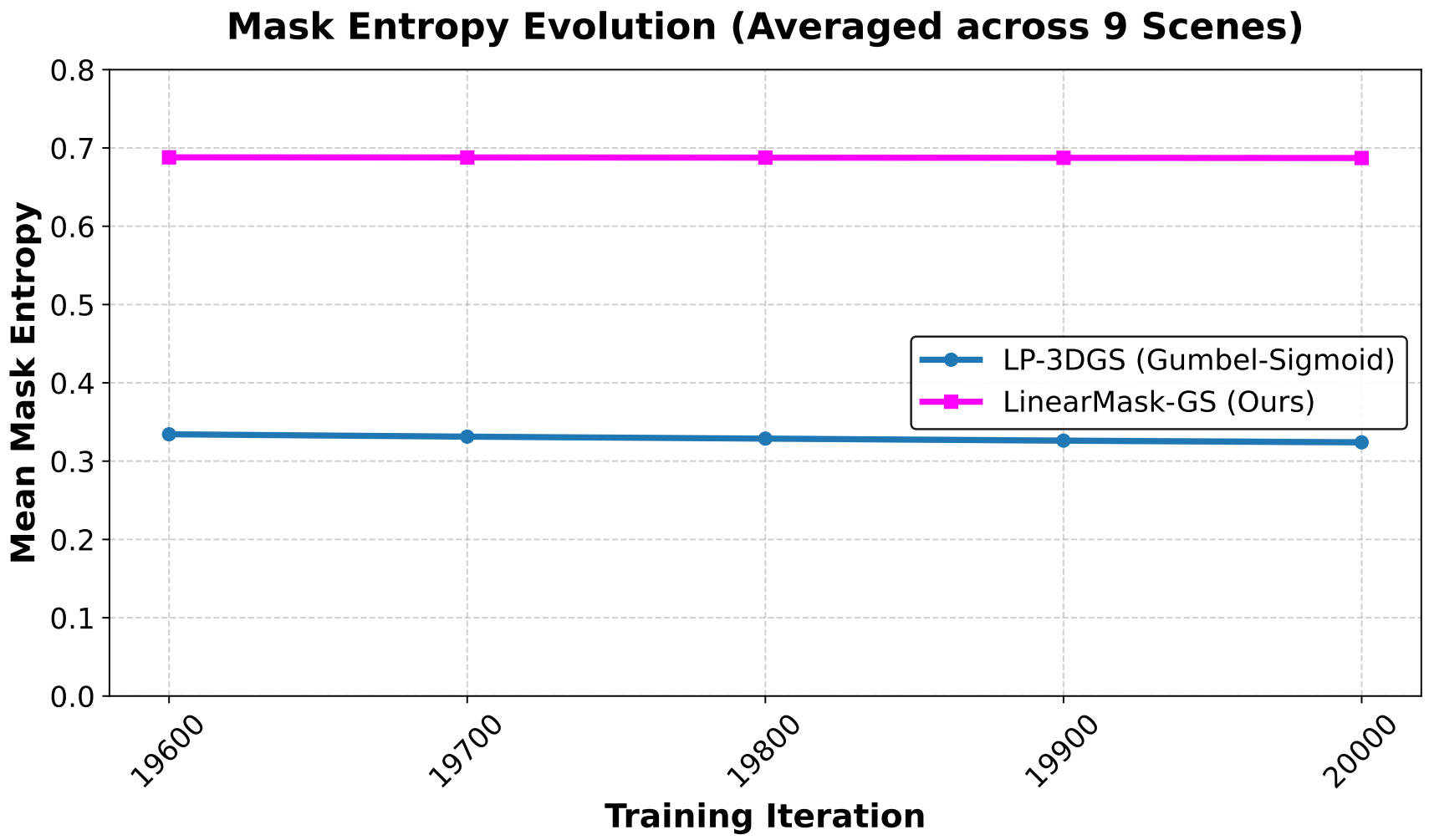


Figure 4: **Mask entropy evolution.** Mean binary entropy of mask values across all nine Mip-NeRF 360 scenes during the measured pruning phase (iterations 19,600–20,000). Shaded regions show standard deviation.

As shown in Figure 4, LP-3DGS exhibits a low entropy ($\approx 0.33$) throughout this phase, confirming that the Gumbel-Sigmoid activation drives an early hard decision. In contrast, LinearMask-GS maintains a substantially higher and stable entropy ($\approx 0.69$), and the narrow standard-deviation band across the nine scenes indicates the robustness of the linear-increment activation. This sustained intermediate regime helps preserve high-frequency details, such as bicycle spokes and foliage, rather than pruning them prematurely.

# 3 Ranking reliability of the learned mask

This section details the protocol behind the ranking-reliability statistics reported in Section 4.4 of the main paper.

**Protocol.** For each of the nine Mip-NeRF 360 scenes we train two runs that are identical except for the mask activation (Gumbel-Sigmoid vs. linear increment), with hard pruning *disabled* so that the mask evolution can be observed without feedback from primitive removal. We snapshot the mask $\psi$ every 100 iterations over the measured part of the mask window (iterations 19,600–20,000) and, after training continues to 30,000 iterations, compute a held-out contribution proxy for every Gaussian: its max-pooled rendering contribution (Eq. 4 of the main paper) at iteration 30k.

**Metrics.** (i) *Decision stability:* at the pruning decision point (20k), we re-sample the Gumbel noise and re-apply the top-$\rho N$ rule; the flip fraction is the share of Gaussians whose keep/prune decision changes between two independent noise draws. (ii) *Saturation:* the fraction of $\psi$ values outside $[0.1, 0.9]$. (iii) *Ranking fidelity:* the Spearman correlation between the ranking induced by $\psi$ and the proxy ranking.

**Results.** Re-sampling the Gumbel noise alone flips 47.3% of keep/prune decisions on average (46–48% on every scene), and the rank correlation between two independent noise draws is below 0.1: the sampled decision is governed by activation noise rather than by a stable learned ranking. The linear mask is deterministic (0% flips). Saturation reaches 52.2% for Gumbel-Sigmoid under this no-pruning protocol vs. 0.00% for linear; the corresponding figure under the standard schedule of the main paper is 51.19% (Section 3.3), i.e., disabling pruning does not alter the picture. The Spearman correlation against the 30k proxy is 0.29 for linear vs. 0.16 for Gumbel-Sigmoid (1.8× higher overall and 2.8× on outdoor scenes). Figure 5 plots both statistics across the mask window; the narrow per-scene spread shows the statistics are scene-invariant rather than driven by a few scenes.

# 4 Effect of mask-window length and placement

A natural question is whether a longer mask-training window would let the Gumbel-Sigmoid mask escape saturation. It does not; it makes saturation worse. We sweep the placement of the mask window on three scenes (*Bicycle*, *Flowers*, *Counter*) by moving its start to 15k, 17.5k, or 19.5k with the end held at the 20k pruning point, so that the sweep simultaneously varies the window *length* (5,000, 2,500, and 500 iterations). Under Gumbel-Sigmoid, the saturated fraction at the decision point rises monotonically as the window lengthens, from 51.1% to 54.8% to 56.2%, while PSNR plateaus (the two longest windows differ by

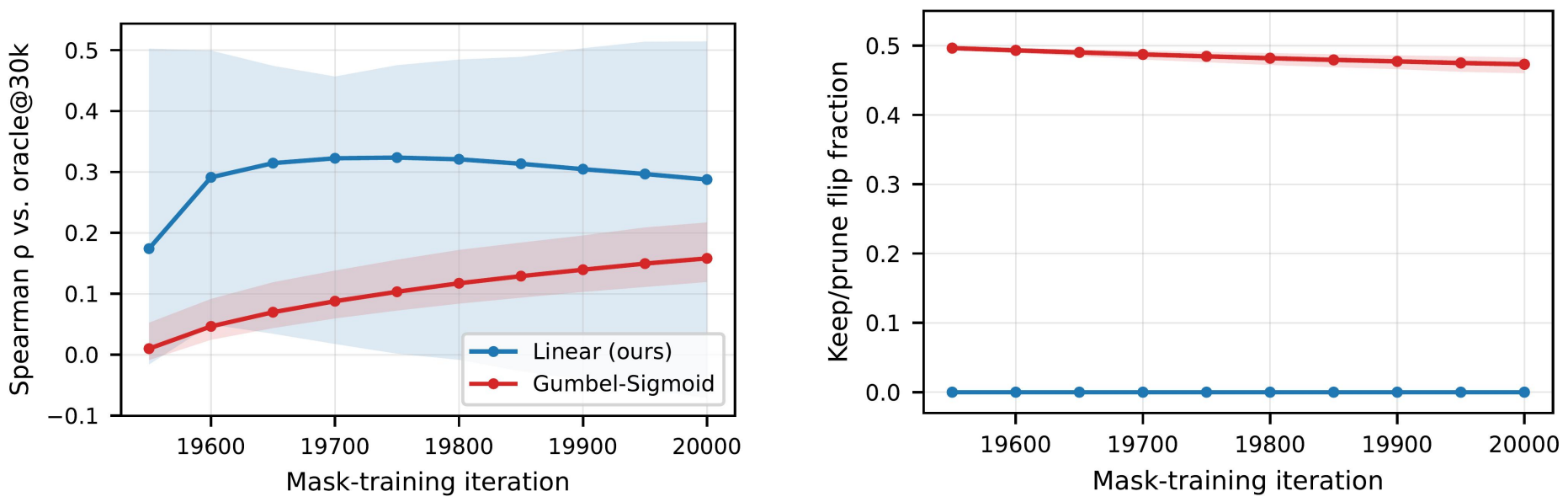


(a) Spearman corr. vs. 30k proxy

(b) Keep/prune flip fraction

Figure 5: Ranking reliability on Mip-NeRF 360 across the mask-training window. (a) Spearman correlation between the mask ranking and the held-out max-pooled contribution proxy at 30k. (b) Fraction of keep/prune decisions flipped by re-sampling the Gumbel activation noise; the linear-increment mask is deterministic. Shaded regions show the min–max range over the nine scenes.

$\pm$0.05 dB). The linear-increment activation is not fully exempt from this trend: its saturated fraction also grows with the window, from 0.0012% to 0.0606% to 0.0782%. It does so, however, at a level at least 700$\times$ below Gumbel-Sigmoid, and in all nine scene–window cells (3 scenes $\times$ 3 window settings) the Gumbel-Sigmoid fraction exceeds the linear one. Its final PSNR varies by at most 0.14 dB across the three placements. Table 1 summarises the sweep.

Table 1: Effect of the mask-window length on saturation, with the pruning point held fixed at 20k. Moving the window start keeps the end at the pruning point, so length and placement vary along a single axis. The saturated fraction is the share of $\psi_i$ outside $[0.1, 0.9]$ at the decision point, averaged over the three scenes (*Bicycle*, *Flowers*, *Counter*). Lengthening the window raises saturation for both activations, but the linear increment remains at least 700$\times$ below Gumbel-Sigmoid at every setting. Quality does not compensate: Gumbel-Sigmoid PSNR differs by $\pm$0.05 dB between the two longest windows, and the linear increment by at most 0.14 dB across all three placements, so window duration is not the binding factor.

| Mask window | | Saturated fraction at 20k (↓) | |
| --- | --- | --- | --- |
| Start | Length | Gumbel-Sigmoid | Linear increment |
| 19.5k | 500 | 51.1% | **0.00%** |
| 17.5k | 2,500 | 54.8% | **0.06%** |
| 15.0k | 5,000 | 56.2% | **0.08%** |

Longer exposure to the steep activation simply gives mask values more time to reach the saturation boundaries; under the fixed-pruning-point protocol, the binding factor is activation steepness, not window duration. The asymmetry mirrors the ranking-reliability analysis of Section 3: a ranking that is stable in time and deterministic in noise is insensitive to when it is read out, whereas a saturating one degrades the longer it saturates.

# 5 Per-Scene Quantitative Metrics on Mip-NeRF 360

Tables 2–6 detail the per-scene performance on the nine Mip-NeRF 360 scenes. Our method consistently matches or outperforms the LP-3DGS pruning baseline in LPIPS, PSNR, and SSIM, while using a drastically smaller number of Gaussians (Table 5) and storage footprint (Table 6). Figure 6 visualizes the same per-scene trade-off: each marker is one scene, and lower-left is better.

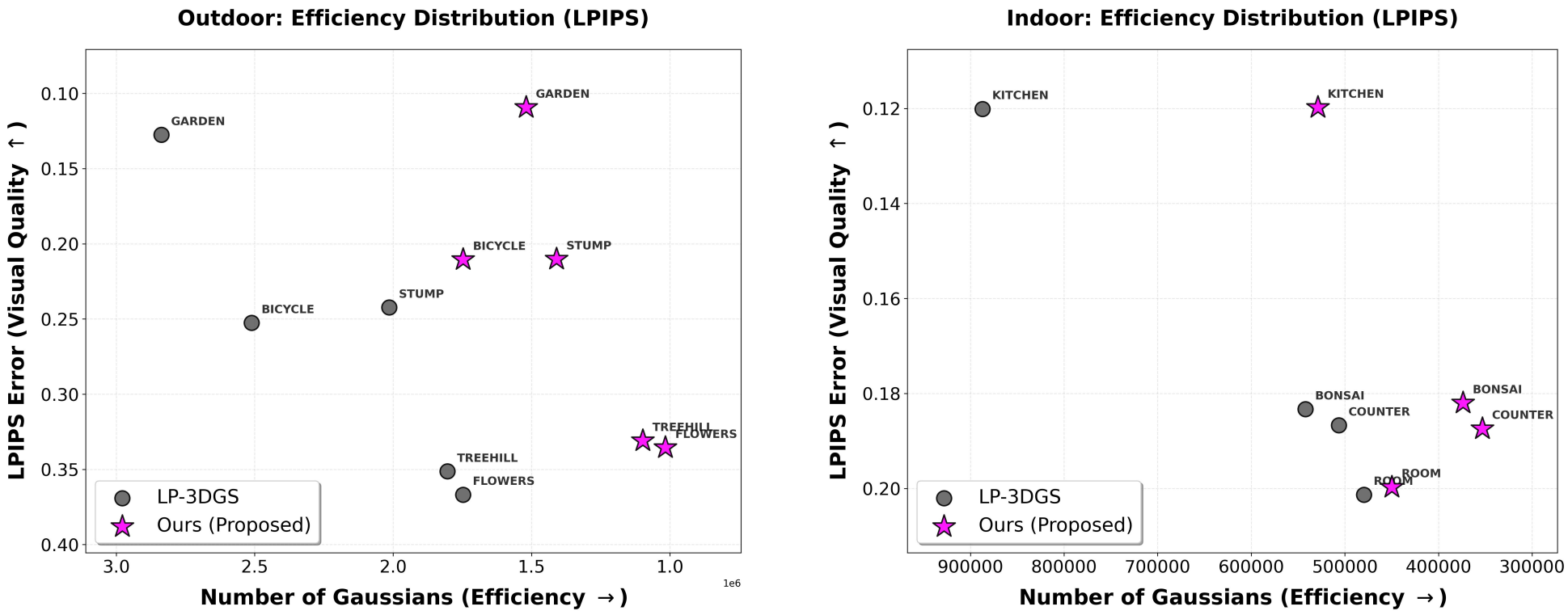


(a) Outdoor scenes (b) Indoor scenes

Figure 6: Per-scene LPIPS vs. number of Gaussians on Mip-NeRF 360. Each marker is one scene; lower-left is better. LinearMask-GS (★) reaches comparable or lower LPIPS with substantially fewer Gaussians than LP-3DGS (●).

Table 2: LPIPS scores (↓) for Mip-NeRF 360 scenes. †Literature rows (Plenoxels, INGP-base, Mip-NeRF 360, 3DGS) report the values given in the original papers; the per-scene entries in those rows are approximate reproductions and do not always average to them. The LP-3DGS and Ours rows are means of our per-scene measurements.

| Method | Bicycle | Bonsai | Counter | Kitchen | Room | Stump | Garden | Flowers | Treehill | Avg. |
|---|---|---|---|---|---|---|---|---|---|---|
| Plenoxels | 0.534 | 0.421 | 0.451 | 0.354 | 0.472 | 0.485 | 0.351 | 0.521 | 0.578 | 0.463† |
| INGP-base | 0.495 | 0.245 | 0.364 | 0.285 | 0.185 | 0.421 | 0.325 | 0.485 | 0.534 | 0.371† |
| Mip-NeRF 360 | 0.354 | 0.198 | 0.211 | 0.154 | 0.225 | 0.285 | 0.175 | 0.395 | 0.364 | 0.237† |
| 3DGS | **0.205** | 0.205 | 0.204 | 0.129 | 0.220 | 0.215 | **0.103** | 0.336 | **0.317** | 0.214† |
| LP-3DGS | 0.253 | 0.183 | 0.187 | 0.120 | 0.201 | 0.242 | 0.128 | 0.367 | 0.351 | 0.226 |
| **Ours** | 0.211 | **0.182** | **0.187** | **0.120** | **0.200** | **0.210** | 0.109 | **0.336** | 0.331 | **0.210** |

# 6 Full quantitative comparison on Tanks & Temples and Deep Blending

Table 7 provides the complete scene-wise comparison on Tanks & Temples and Deep Blending that was summarized in Section 4.2 of the main paper. We extend the comparison with recent fast-training and compact 3DGS baselines, using their scene-wise paper results when available. The final two rows report the main LinearMask-GS result and the FastGS-backbone variant LinearMask-GS★.

Table 3: PSNR scores (↑) for Mip-NeRF 360 scenes. †Literature rows (Plenoxels, INGP-base, Mip-NeRF 360, 3DGS) report the values given in the original papers; the per-scene entries in those rows are approximate reproductions and do not always average to them. The LP-3DGS and Ours rows are means of our per-scene measurements.

| Method | Bicycle | Bonsai | Counter | Kitchen | Room | Stump | Garden | Flowers | Treehill | Avg. |
|---|---|---|---|---|---|---|---|---|---|---|
| Plenoxels | 21.73 | 25.54 | 24.85 | 24.71 | 22.51 | 23.23 | 23.51 | 20.07 | 21.54 | 23.08† |
| INGP-base | 22.17 | 28.31 | 26.85 | 27.54 | 30.54 | 24.12 | 24.96 | 21.14 | 22.05 | 25.30† |
| Mip-NeRF 360 | 24.37 | **33.46** | **29.55** | **32.23** | 31.63 | 26.40 | 26.98 | 21.52 | 22.49 | 27.69† |
| 3DGS | 25.25 | 31.13 | 28.70 | 30.32 | 30.63 | 26.55 | 27.41 | 21.52 | 22.49 | 27.21† |
| LP-3DGS | 25.10 | 32.09 | 28.94 | 31.52 | 31.49 | 26.69 | 27.29 | 21.38 | 22.71 | 27.47 |
| **Ours** | **25.67** | 32.02 | 29.05 | 31.50 | **31.70** | **26.94** | **27.75** | **21.86** | **22.85** | **27.70** |

Table 4: SSIM scores (↑) for Mip-NeRF 360 scenes. †Literature rows (Plenoxels, INGP-base, Mip-NeRF 360, 3DGS) report the values given in the original papers; the per-scene entries in those rows are approximate reproductions and do not always average to them. The LP-3DGS and Ours rows are means of our per-scene measurements.

| Method | Bicycle | Bonsai | Counter | Kitchen | Room | Stump | Garden | Flowers | Treehill | Avg. |
|---|---|---|---|---|---|---|---|---|---|---|
| Plenoxels | 0.447 | 0.864 | 0.795 | 0.782 | 0.878 | 0.498 | 0.701 | 0.351 | 0.315 | 0.626† |
| INGP-base | 0.495 | 0.912 | 0.854 | 0.831 | 0.899 | 0.542 | 0.723 | 0.431 | 0.378 | 0.671† |
| Mip-NeRF 360 | 0.685 | 0.941 | 0.894 | 0.920 | 0.913 | 0.744 | 0.813 | 0.583 | 0.632 | 0.792† |
| 3DGS | 0.771 | 0.938 | 0.905 | 0.922 | 0.913 | 0.775 | 0.868 | 0.605 | 0.638 | 0.815† |
| LP-3DGS | 0.744 | **0.946** | **0.914** | 0.931 | **0.926** | 0.771 | 0.855 | 0.584 | 0.636 | 0.812 |
| **Ours** | **0.779** | 0.944 | 0.913 | **0.931** | 0.925 | **0.787** | **0.872** | **0.623** | **0.655** | **0.825** |

Table 5: Number of final Gaussians (millions, ↓) for Mip-NeRF 360 scenes. †3DGS values are those reported in the original paper; the per-scene entries in this row are approximate reproductions and do not average to it. The LP-3DGS and Ours rows are means of our per-scene measurements.

| Method | Bicycle | Bonsai | Counter | Kitchen | Room | Stump | Garden | Flowers | Treehill | Avg. |
|---|---|---|---|---|---|---|---|---|---|---|
| 3DGS | 5.7M | 1.3M | 1.2M | 1.8M | 1.5M | 3.3M | 5.8M | 3.6M | 3.8M | 3.36M† |
| LP-3DGS | 2.51M | 0.54M | 0.51M | 0.88M | 0.48M | 2.01M | 2.83M | 1.75M | 1.80M | 1.48M |
| **Ours** | **1.75M** | **0.37M** | **0.35M** | **0.53M** | **0.45M** | **1.41M** | **1.52M** | **1.02M** | **1.10M** | **0.94M** |

Table 6: Storage size (MB, ↓) for Mip-NeRF 360 scenes. †3DGS values are those reported in the original paper; the per-scene entries in this row are approximate reproductions and do not average to it. The LP-3DGS and Ours rows are means of our per-scene measurements.

| Method | Bicycle | Bonsai | Counter | Kitchen | Room | Stump | Garden | Flowers | Treehill | Avg. |
|---|---|---|---|---|---|---|---|---|---|---|
| 3DGS | 1450.2 | 340.8 | 320.5 | 480.2 | 390.4 | 1050.2 | 1280.5 | 750.4 | 850.1 | 734.0† |
| LP-3DGS | 557.9 | 101.6 | 95.2 | 175.0 | 90.4 | 471.7 | 568.6 | 369.0 | 397.7 | 314.1 |
| **Ours** | **432.1** | **88.4** | **83.5** | **125.1** | **86.4** | **333.5** | **395.4** | **240.4** | **256.0** | **226.8** |

# 7 Per-scene qualitative comparison (remaining six scenes)

Figure 7 extends Figure 2 of the main paper with per-pixel absolute error maps for the six Mip-NeRF 360 scenes that were not shown there: *Garden*, *Stump*, *Treehill* (outdoor), and *Bonsai*, *Kitchen*, *Room* (indoor). The visualization conventions are identical to the main paper's Figure 2: per-image PSNR is reported below each rendered image and per-image MAE below each error map, and each scene uses its own blue-to-red colormap from low to high absolute error. Across all six scenes, LinearMask-GS produces lower MAE and fewer

saturated high-error (red) regions than both LP-3DGS and 3DGS.

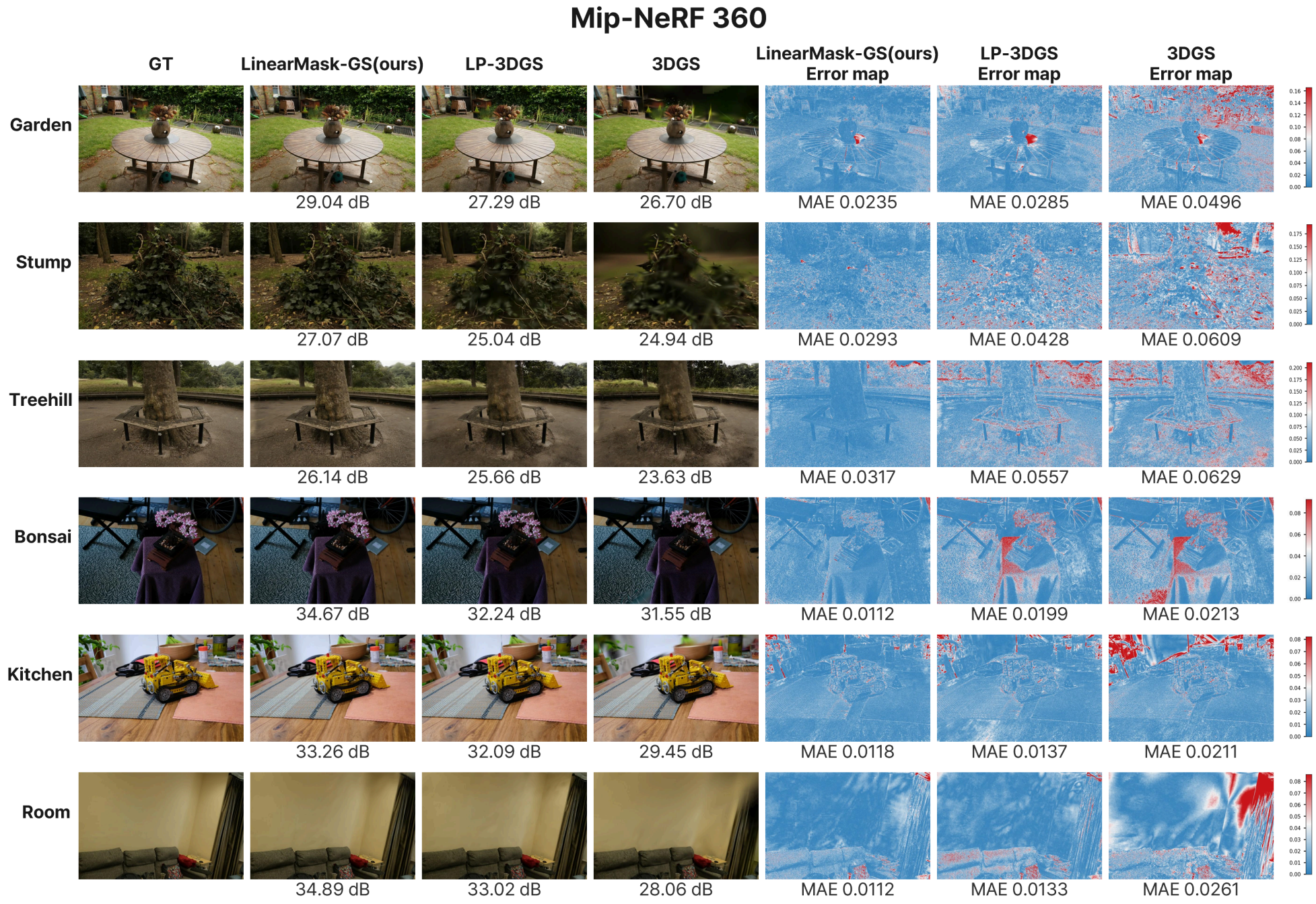


Figure 7: Per-pixel qualitative comparison on the six Mip-NeRF 360 scenes not shown in the main paper's Figure 2. Each row shows one scene; columns from left to right show the ground truth, the rendered images from LinearMask-GS (ours), LP-3DGS, and 3DGS, followed by per-pixel absolute error maps for the same three methods. Per-image PSNR (dB) is reported below each render and per-image MAE below each error map; error maps use a scene-specific blue-to-red colormap (low to high absolute error).

# 8 Minimum-survival ratio $\rho$ sweep for the linear-increment activation

Figure 8 shows the per-scene LPIPS curves against the number of Gaussians for $\rho \in \{0.20, 0.30, 0.40\}$, and Table 8 extends the sweep to $\rho \in \{0.10, 0.20, 0.30, 0.40, 0.50\}$ with overall, outdoor, and indoor averages, mirroring the format of Table 9. Under the linear-increment activation, $\psi_i$ is unimodal and its relative order is preserved rather than collapsed onto the saturation points (Section 3.3 of the main paper), so the top-$\rho N$ rule selects a consistent, importance-ordered subset whose size is set by $\rho$.

Across every scene, LPIPS decreases monotonically as $\rho$ increases, at the expected cost of more retained Gaussians. The returns are asymmetric about a knee at $\rho$=0.30: $\rho$=0.20 produces a visible LPIPS increase on most scenes (−0.28 dB PSNR on aggregate), indicating that this regime removes too many geometrically meaningful primitives, whereas $\rho$=0.40 retains 34% more Gaussians for only +0.03 dB and a 0.006 LPIPS gain. We therefore use $\rho$=0.30 as the default operating point. We emphasize that the linear-increment activation

does not exempt the trade-off from saturation: beyond $\rho$=0.30 its aggregate PSNR gains total only +0.04 dB up to $\rho$=0.50, just as the Gumbel-Sigmoid baseline saturates over the same range. The benefit of the stable ranking lies not in the *shape* of this curve but in its *position*, which Section 9 quantifies.

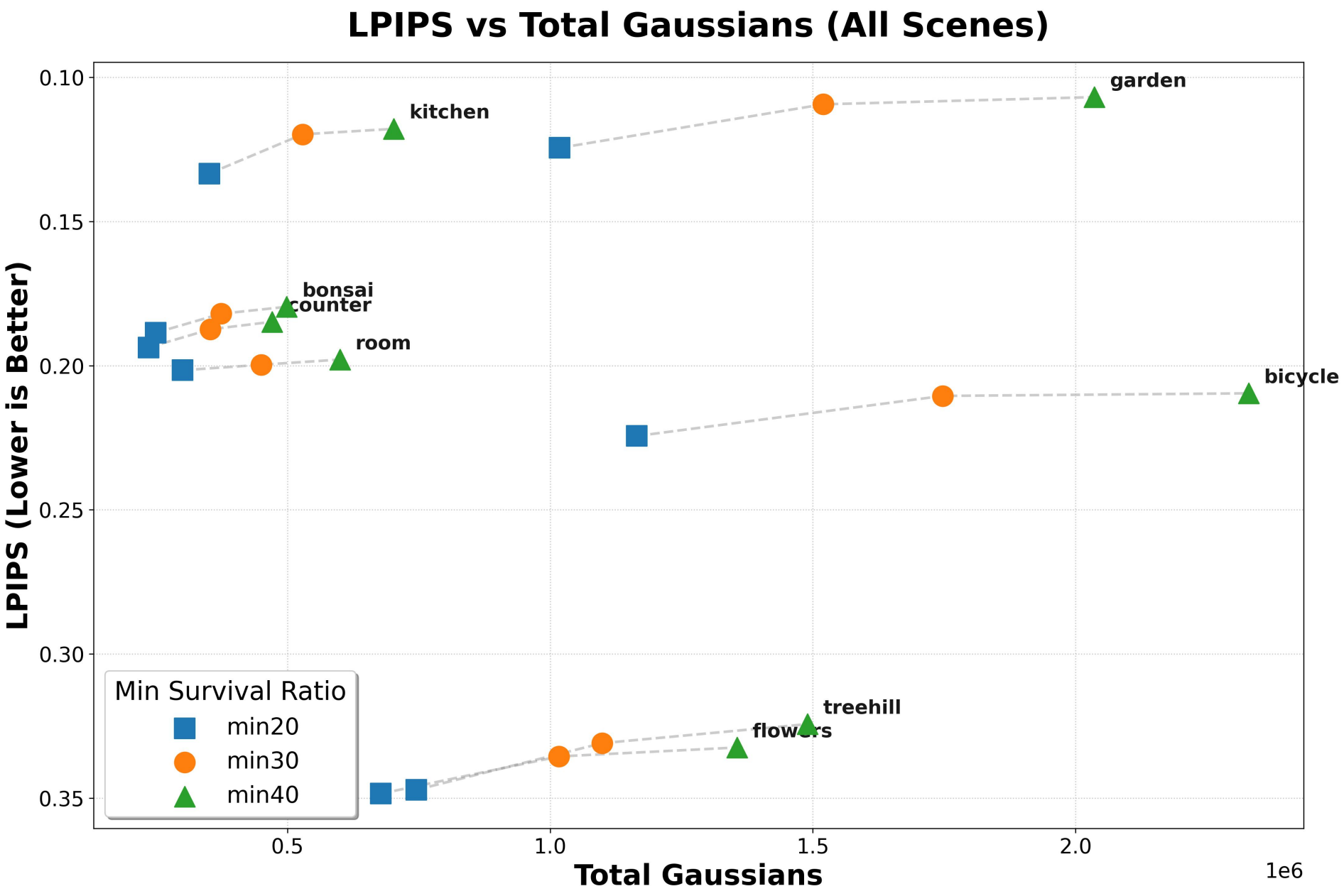


Figure 8: Effect of the minimum-survival ratio $\rho$ on Mip-NeRF 360: per-scene LPIPS vs. number of Gaussians for $\rho \in \{0.20, 0.30, 0.40\}$. Lower LPIPS with fewer Gaussians indicates a better quality–compression trade-off. Increasing $\rho$ monotonically reduces LPIPS at the cost of more retained Gaussians; $\rho$=0.20 already produces a visible LPIPS increase on most scenes, while $\rho$=0.40 retains significantly more primitives without proportional perceptual gains. We use $\rho$=0.30 as the default operating point.

# 9 $\rho$ sweep: linear increment vs. the matched Gumbel-Sigmoid baseline

Table 9 provides the analogous $\rho$ sweep for the matched Gumbel-Sigmoid baseline (i.e., LP-3DGS with max-pooling aggregation, the "Max + Gumbel-Sigmoid" row of Table 4 in the main paper), under the identical top-$\rho N$ rule and schedule as Table 8. Because top-$\rho N$ retains the same fraction of a near-identical primitive pool, equal $\rho$ implies matched budgets: #G differs by at most 0.01M at every $\rho$, so each row of the two tables is a same-budget comparison.

Both curves saturate beyond $\rho$=0.30. For Gumbel-Sigmoid, PSNR rises by +0.49 dB from $\rho$=0.10 to 0.20 and +0.24 dB from 0.20 to 0.30, then by only +0.11 and +0.00 dB; for the linear increment, the corresponding steps are +0.48, +0.28, +0.03, and +0.01 dB. The

*shape* of the quality–compression trade-off is therefore not what separates the two activations: both follow the diminishing returns of the scene's importance spectrum, with retained Gaussians scaling nearly linearly in $\rho$ while quality gains collapse past the knee.

What separates them is *position*. At every $\rho$ in the sweep, the linear-increment activation is +0.40 to +0.48 dB PSNR above the Gumbel-Sigmoid baseline at the same budget, and the baseline cannot match the linear curve anywhere: its best point (27.33 dB / 0.2321 LPIPS at $\rho$=0.50 with 1.57M Gaussians) falls below even the linear $\rho$=0.20 setting (27.42 dB / 0.2194 LPIPS at 0.63M, i.e., 2.5× fewer primitives), and it does not reach the linear default (27.70 dB / 0.2095 at $\rho$=0.30 with 0.94M; Table 1 of the main paper) despite using 67% more primitives. The outdoor-vs.-indoor split shows the offset is broad rather than scene-specific: at every $\rho$ in this sweep, the Gumbel-Sigmoid baseline lags the linear increment in both subsets on all three metrics at every $\rho$ (Table 8): the outdoor PSNR margin ranges from +0.48 to +0.55 dB and the indoor margin from +0.29 to +0.44 dB, with SSIM and LPIPS showing the same sign throughout.

These same-budget deficits are the downstream signature of the ranking analysis in Section 3: a saturated, noise-driven mask still produces a $\rho$-curve of the ordinary shape, but the subset it selects at each budget is less aligned with importance, and this appears as a near-uniform vertical offset of the entire curve rather than as a change in its slope.

# 10 Direct pruning by the raw max-pooled score

To isolate the contribution of the learned mask beyond its ingredients, we prune directly by the raw max-pooled importance score: the schedule is identical to LinearMask-GS, but at iteration 20k the top-30% of Gaussians by $S_i$ (Eq. 4 of the main paper) are retained with no learned mask, followed by the same fine-tuning. On the nine Mip-NeRF 360 scenes this mask-free baseline reaches 27.5725 dB / 0.8231 / 0.2143 at 0.9612M, while the learned mask reaches higher PSNR on eight of the nine scenes at a smaller budget (27.7044 dB / 0.8254 / 0.2095 at 0.9444M), a +0.1319 dB gain. The sole exception is *Treehill* (−0.17 dB), the scene with the largest sky and distant-vegetation area, where the raw score already ranks primitives well. Table 10 reports the per-scene PSNR.

# 11 Budget-matched comparison with Speedy-Splat

Table 1 of the main paper compares methods at their native operating points, where Speedy-Splat uses 0.30M Gaussians. For a comparison at a matched primitive budget, we retrain LinearMask-GS★ (FastGS backbone) under primitive budgets of 0.25M, 0.30M, and 0.35M. Averaged over the nine Mip-NeRF 360 scenes and three independent seeds at 0.30M, LinearMask-GS★ reaches 27.40 dB / 0.798 / 0.230, vs. 26.91 / 0.781 / 0.295 for Speedy-Splat, i.e., better on all three metrics at the identical budget. It remains ahead even at the smaller 0.25M budget, and improves further at 0.35M (Table 11).

# 12 Extended cross-backbone evaluation

Table 3 of the main paper evaluates DropGaussian on a single LLFF scene (*fern*) and Octree-GS on a single Bungee-NeRF scene (*amsterdam*). Here we extend both.

**DropGaussian: five scenes, three runs.** We evaluate five LLFF scenes (9-view; *fern*, *flower*, *fortress*, *horns*, and *room*) with the masked model repeated over three independent training runs (the public DropGaussian code does not expose an explicit seed flag, so repetitions correspond to independent launches). Averaged over the three runs, LinearMask-GS improves PSNR on every scene, by +0.31 to +0.52 dB (mean +0.42 dB) over the unpruned backbone, while retaining 41% fewer primitives than the Gumbel variant (Table 12). The single *fern* run reported in Table 3 of the main paper gives a smaller gain (+0.03 dB), indicating that single runs of this sparse-view backbone are noisy and motivating the repeated protocol used here. Table 12 reports the per-scene, per-run results.

**Octree-GS: two additional scenes.** Beyond *amsterdam* (a 2.9× anchor reduction at +0.12 dB; Table 3 of the main paper), we evaluate the *bicycle* and *garden* scenes of Mip-NeRF 360: anchors drop by 40.2% relative to the Gumbel variant at a mean PSNR change of −0.11 dB. We report the two scene sets separately rather than averaging them, since the sign of the (small) PSNR change differs; across all three scenes, quality moves by at most ≈0.1 dB in either direction while the anchor count is reduced by 40% or more.

# 13 Why mask learning is so fast on the FastGS backbone

As reported in Table 2 of the main paper, applying our mask on the FastGS backbone (LinearMask-GS★) does not increase, and in fact reduces, the total training time, whereas the 3DGS-backbone variant adds a mask-training stage on top of the base pipeline. This section explains why. We first decompose the mask-learning speed gap (Table 14), then account for the net change in total training time over the 20k–30k iteration range (Table 15).

**Mask-learning speed gap.** The mask-learning procedure that takes 110 s on the 3DGS backbone completes in only 4.6 s on the FastGS backbone; both are measured on the same RTX 4090, so the gap reflects the backbone and rasterizer rather than any hardware difference. Table 14 attributes this 24× gap to two directly measured factors. The dominant factor is the number of Gaussians present when the mask window begins (Factor 1, 6.36×): averaged over the nine Mip-NeRF 360 scenes, the 3DGS backbone enters the mask window with 3.14 M Gaussians, whereas FastGS has already reduced this to 0.493 M through its cumulative pruning events, so each mask-training iteration rasterizes about 6.36× fewer primitives. The second factor is the FastGS rasterizer itself (Factor 2, 3.7×), which processes far fewer tiles per splat. The two factors combine to $6.36 \times 3.7 \approx 24\times$, consistent with the directly measured times since $110\,\text{s}/24 \approx 4.6\,\text{s}$.

**Net effect on total training time.** Although LinearMask-GS★ adds a 4.6 s mask-learning stage that FastGS does not have, it more than recovers this cost during the rest of training. Table 15 reports the wall-clock time over the 20k–30k iteration range, averaged across the nine Mip-NeRF 360 scenes. Because LinearMask-GS★ prunes additional primitives, it renders fewer Gaussians per iteration and completes this range in 28.3 s versus 39.2 s for FastGS native, a 10.9 s saving. Subtracting the 4.6 s mask-learning cost still leaves a net saving of 6.3 s per scene, which is why the total training time in Table 2 of the main paper decreases from 1.93 to 1.81 minutes rather than increasing.

In short, the speed advantage of LinearMask-GS$^\star$ comes entirely from operating on a backbone that has already pruned most primitives before mask learning begins, combined with a faster rasterizer, and not from any change to the masking formulation itself, which is identical across backbones.

# 14 Supplementary Video

Alongside this document, we provide a supplementary video archive containing split-screen comparison renderings for all 13 evaluated scenes (9 from Mip-NeRF 360, 2 from Deep Blending, and 2 from Tanks & Temples). Each video compares rendered test views in a $2 \times 2$ grid: ground truth (top-left), 3DGS (top-right), LP-3DGS (bottom-left), and LinearMask-GS (bottom-right). Each panel displays the total number of Gaussians used. Despite using substantially fewer Gaussians, our method preserves perceptual quality comparable to the original 3DGS and superior to LP-3DGS, retaining high-frequency details.

Table 7: Quantitative comparison on Tanks & Temples and Deep Blending. Best values are in bold. Fast-training baseline rows use the scene-wise values reported by FastGS [9]; ★ marks our FastGS-backbone run that substitutes the VCP module with LinearMask-GS.

| Method | Metric | Tanks & Temples | | | Deep Blending | | |
|---|---|---|---|---|---|---|---|
| | | Truck | Train | **Avg** | Dr Johnson | Playroom | **Avg** |
| Plenoxels [4] | PSNR↑ | 23.221 | 18.927 | 21.074 | 23.142 | 22.980 | 23.061 |
| | SSIM↑ | 0.774 | 0.663 | 0.718 | 0.787 | 0.802 | 0.794 |
| | LPIPS↓ | 0.335 | 0.422 | 0.378 | 0.521 | 0.499 | 0.510 |
| INGP-Base [8] | PSNR↑ | 23.260 | 20.170 | 21.715 | 27.750 | 19.483 | 23.616 |
| | SSIM↑ | 0.779 | 0.666 | 0.722 | 0.839 | 0.754 | 0.796 |
| | LPIPS↓ | 0.274 | 0.386 | 0.330 | 0.381 | 0.465 | 0.423 |
| INGP-Big [8] | PSNR↑ | 23.383 | 20.456 | 21.919 | 28.257 | 21.665 | 24.961 |
| | SSIM↑ | 0.800 | 0.689 | 0.744 | 0.854 | 0.779 | 0.816 |
| | LPIPS↓ | 0.249 | 0.360 | 0.304 | 0.352 | 0.428 | 0.390 |
| Mip-NeRF 360 [1] | PSNR↑ | 24.912 | 19.523 | 22.217 | 29.140 | 29.657 | 29.398 |
| | SSIM↑ | 0.857 | 0.660 | 0.758 | 0.901 | 0.900 | 0.900 |
| | LPIPS↓ | 0.159 | 0.354 | 0.256 | **0.237** | 0.252 | 0.244 |
| 3DGS [6] | PSNR↑ | 25.187 | 21.097 | 23.142 | 28.766 | 30.044 | 29.405 |
| | SSIM↑ | 0.879 | 0.802 | 0.840 | 0.899 | 0.906 | 0.902 |
| | LPIPS↓ | 0.148 | 0.218 | 0.183 | 0.244 | **0.241** | **0.242** |
| | #G (M)↓ | 1.09 | 2.59 | 1.84 | 3.31 | 2.33 | 2.82 |
| LP-3DGS [10] | PSNR↑ | 25.376 | 21.822 | 23.599 | 29.223 | 30.154 | 29.690 |
| | SSIM↑ | 0.877 | 0.807 | 0.842 | 0.901 | 0.908 | 0.905 |
| | LPIPS↓ | 0.154 | 0.222 | 0.188 | 0.248 | 0.248 | 0.248 |
| | #G (M)↓ | 1.33 | 0.58 | 0.96 | 1.65 | 1.15 | 1.40 |
| Mini-Splatting [3] | PSNR↑ | 25.320 | 21.600 | 23.460 | 29.510 | 30.470 | 29.990 |
| | SSIM↑ | 0.879 | 0.809 | 0.844 | **0.905** | 0.908 | **0.907** |
| | LPIPS↓ | **0.139** | 0.223 | 0.181 | 0.246 | **0.241** | 0.244 |
| | #G (M)↓ | 0.32 | 0.28 | 0.30 | 0.60 | 0.51 | 0.56 |
| Speedy-Splat [5] | PSNR↑ | 25.180 | 21.590 | 23.380 | 29.070 | 29.770 | 29.420 |
| | SSIM↑ | 0.863 | 0.768 | 0.816 | 0.898 | 0.898 | 0.898 |
| | LPIPS↓ | 0.192 | 0.292 | 0.242 | 0.269 | 0.274 | 0.272 |
| | #G (M)↓ | 0.26 | **0.11** | **0.18** | 0.31 | **0.18** | 0.25 |
| Taming-3DGS [7] | PSNR↑ | 25.270 | 22.500 | 23.890 | 29.040 | 29.960 | 29.500 |
| | SSIM↑ | 0.865 | 0.802 | 0.833 | 0.888 | 0.901 | 0.894 |
| | LPIPS↓ | 0.187 | 0.240 | 0.214 | 0.292 | 0.264 | 0.278 |
| | #G (M)↓ | 0.27 | 0.37 | 0.32 | **0.19** | 0.40 | 0.29 |
| DashGaussian [2] | PSNR↑ | **25.800** | 22.190 | 24.000 | 29.130 | 30.170 | 29.650 |
| | SSIM↑ | **0.886** | **0.819** | **0.853** | 0.903 | **0.909** | 0.906 |
| | LPIPS↓ | 0.150 | **0.206** | **0.178** | 0.250 | 0.243 | 0.246 |
| | #G (M)↓ | 1.43 | 1.00 | 1.21 | 1.51 | 2.38 | 1.94 |
| FastGS [9] | PSNR↑ | 25.730 | **22.570** | 24.150 | 29.500 | 30.570 | 30.030 |
| | SSIM↑ | 0.872 | 0.805 | 0.839 | 0.898 | 0.905 | 0.901 |
| | LPIPS↓ | 0.178 | 0.242 | 0.210 | 0.275 | 0.266 | 0.270 |
| | #G (M)↓ | **0.25** | 0.23 | 0.24 | 0.25 | 0.19 | **0.22** |
| **LinearMask-GS** | PSNR↑ | 25.793 | 22.546 | 24.170 | 29.602 | **30.583** | 30.093 |
| | SSIM↑ | 0.882 | 0.811 | 0.846 | 0.903 | **0.909** | 0.906 |
| | LPIPS↓ | 0.151 | 0.223 | 0.187 | 0.245 | 0.245 | 0.245 |
| | #G (M)↓ | 0.78 | 0.33 | 0.55 | 0.99 | 0.70 | 0.85 |
| **LinearMask-GS★** | PSNR↑ | 25.787 | 22.565 | **24.176** | **29.672** | 30.555 | **30.113** |
| | SSIM↑ | 0.861 | 0.790 | 0.825 | 0.900 | 0.898 | 0.899 |
| | LPIPS↓ | 0.172 | 0.236 | 0.204 | 0.264 | 0.270 | 0.267 |
| | #G (M)↓ | 0.31 | 0.26 | 0.29 | 0.33 | 0.25 | 0.29 |

Table 8: $\rho$ sweep for the linear-increment activation on Mip-NeRF 360 (top-$\rho N$ rule, identical schedule to Table 9). Overall averages are over all nine scenes; outdoor averages are over five scenes (*Bicycle*, *Flowers*, *Garden*, *Stump*, *Treehill*) and indoor averages over four scenes (*Bonsai*, *Counter*, *Kitchen*, *Room*). #G in millions of Gaussians. Overall SSIM/LPIPS are the subset averages recombined as $(5\,\text{out} + 4\,\text{in})/9$ and may differ from the independently rounded values of Table 1 of the main paper in the last digit. The default $\rho$=0.30 row matches Table 1 of the main paper.

| $\rho$ | Overall (9 scenes) | | | | Outdoor (5 scenes) | | | | Indoor (4 scenes) | | | |
|---|---|---|---|---|---|---|---|---|---|---|---|---|
| | PSNR↑ | SSIM↑ | LPIPS↓ | #G (M)↓ | PSNR↑ | SSIM↑ | LPIPS↓ | #G (M)↓ | PSNR↑ | SSIM↑ | LPIPS↓ | #G (M)↓ |
| 0.10 | 26.94 | 0.7951 | 0.2478 | 0.31 | 24.58 | 0.7080 | 0.2780 | 0.45 | 29.89 | 0.9040 | 0.2100 | 0.14 |
| 0.20 | 27.42 | 0.8127 | 0.2194 | 0.63 | 24.81 | 0.7260 | 0.2510 | 0.91 | 30.68 | 0.9210 | 0.1800 | 0.28 |
| 0.30 | 27.70 | 0.8254 | 0.2095 | 0.94 | 25.01 | 0.7432 | 0.2394 | 1.36 | 31.07 | 0.9283 | 0.1723 | 0.43 |
| 0.40 | 27.73 | 0.8267 | 0.2031 | 1.26 | 25.02 | 0.7440 | 0.2320 | 1.81 | 31.11 | 0.9300 | 0.1670 | 0.57 |
| 0.50 | 27.74 | 0.8271 | 0.2012 | 1.57 | 25.03 | 0.7440 | 0.2310 | 2.26 | 31.12 | 0.9310 | 0.1640 | 0.71 |

Table 9: $\rho$ sweep for the matched Gumbel-Sigmoid baseline on Mip-NeRF 360. Overall averages are over all nine scenes; outdoor averages are over five scenes (*Bicycle*, *Flowers*, *Garden*, *Stump*, *Treehill*) and indoor averages over four scenes (*Bonsai*, *Counter*, *Kitchen*, *Room*). #G in millions of Gaussians. As $\rho$ grows, PSNR/LPIPS gains shrink rapidly for both activations while #G scales nearly linearly; even at $\rho$=0.50, this baseline does not reach the LinearMask-GS default (27.70 dB at 0.94M, $\rho$=0.30; see Table 1 of the main paper and Table 8).

| $\rho$ | Overall (9 scenes) | | | | Outdoor (5 scenes) | | | | Indoor (4 scenes) | | | |
|---|---|---|---|---|---|---|---|---|---|---|---|---|
| | PSNR↑ | SSIM↑ | LPIPS↓ | #G (M)↓ | PSNR↑ | SSIM↑ | LPIPS↓ | #G (M)↓ | PSNR↑ | SSIM↑ | LPIPS↓ | #G (M)↓ |
| 0.10 | 26.49 | 0.7773 | 0.2929 | 0.32 | 24.03 | 0.6775 | 0.3456 | 0.46 | 29.57 | 0.9021 | 0.2270 | 0.14 |
| 0.20 | 26.98 | 0.7968 | 0.2590 | 0.64 | 24.33 | 0.7011 | 0.3070 | 0.92 | 30.29 | 0.9164 | 0.1991 | 0.28 |
| 0.30 | 27.22 | 0.8037 | 0.2449 | 0.95 | 24.49 | 0.7093 | 0.2903 | 1.37 | 30.63 | 0.9217 | 0.1881 | 0.43 |
| 0.40 | 27.33 | 0.8073 | 0.2370 | 1.26 | 24.53 | 0.7136 | 0.2812 | 1.82 | 30.82 | 0.9244 | 0.1819 | 0.57 |
| 0.50 | 27.33 | 0.8091 | 0.2321 | 1.57 | 24.55 | 0.7158 | 0.2751 | 2.26 | 30.82 | 0.9257 | 0.1783 | 0.71 |

Table 10: Per-scene PSNR (dB, ↑) of direct top-30% pruning by the raw max-pooled score vs. LinearMask-GS on Mip-NeRF 360. Identical schedule and fine-tuning; the only difference is the learned mask. The learned mask wins on eight of nine scenes, the exception being *Treehill*.

| Method | Bicycle | Bonsai | Counter | Kitchen | Room | Stump | Garden | Flowers | Treehill | Avg. |
|---|---|---|---|---|---|---|---|---|---|---|
| Direct top-30% by $S_i$ | 25.4887 | 31.8328 | 28.8942 | 31.3281 | 31.5274 | 26.7869 | 27.5812 | 21.6934 | **23.0198** | 27.5725 |
| **LinearMask-GS (Ours)** | **25.6700** | **32.0200** | **29.0500** | **31.5000** | **31.7000** | **26.9400** | **27.7500** | **21.8600** | 22.8500 | **27.7044** |
| Δ | +0.1813 | +0.1872 | +0.1558 | +0.1719 | +0.1726 | +0.1531 | +0.1688 | +0.1666 | −0.1698 | +0.1319 |

Table 11: Budget-matched comparison with Speedy-Splat on Mip-NeRF 360, averaged over nine scenes. LinearMask-GS$^{\star}$ rows use the FastGS backbone; all rows are means over three independent seeds. At the 0.30M budget the three seeds give 27.38, 27.40 and 27.41 dB, a spread of 0.03 dB, so even the weakest seed leads Speedy-Splat by 0.47 dB.

| Method | #G↓ | PSNR↑ | SSIM↑ | LPIPS↓ |
|---|---|---|---|---|
| Speedy-Splat | 0.30M | 26.91 | 0.781 | 0.295 |
| Ours | 0.25M | 27.06 | 0.791 | 0.241 |
| Ours | 0.30M | 27.40 | 0.798 | 0.230 |
| Ours | 0.35M | **27.60** | **0.802** | **0.224** |

Table 12: DropGaussian + LinearMask-GS on five LLFF scenes (9-view). PSNR values for the masked model are means over three independent training runs (the public code exposes no seed flag, so repetitions correspond to independent launches); the unpruned and Gumbel-variant reference columns are the single-run values also used in Table 3 of the main paper. Note the two reference points: ΔPSNR is measured against the *unpruned* backbone, whereas the reduction is measured against the *Gumbel* variant, following the convention of Section 4.4 of the main paper.

| Scene | PSNR↑ (dB) | | | #Primitives↓ | | |
|---|---|---|---|---|---|---|
| | No pruning | + LinearMask-GS | Δ | + LP-3DGS (Gumbel) | + LinearMask-GS | Reduction |
| *fern* | 26.6495 | 26.9623 | +0.3128 | 62,182 | 36,656 | 41.05% |
| *flower* | 27.8341 | 28.2594 | +0.4253 | 82,070 | 48,233 | 41.23% |
| *fortress* | 31.2183 | 31.7365 | +0.5182 | 72,125 | 42,431 | 41.17% |
| *horns* | 27.9452 | 28.4328 | +0.4876 | 90,304 | 53,379 | 40.89% |
| *room* | 32.6074 | 32.9710 | +0.3636 | 57,442 | 33,741 | 41.26% |
| **Mean** | 29.2509 | **29.6724** | **+0.4215** | 72,825 | **42,888** | **41.12**% |

Table 13: Octree-GS + LinearMask-GS per-scene results. *Amsterdam* is the Bungee-NeRF scene of Table 3 of the main paper; *Bicycle* and *Garden* are the two added Mip-NeRF 360 scenes. As in Table 12, ΔPSNR is measured against the unpruned backbone and the reduction against the Gumbel variant.

| Scene | PSNR↑ (dB) | | | #Anchors↓ | | |
|---|---|---|---|---|---|---|
| | No pruning | + LinearMask-GS | Δ | + LP-3DGS (Gumbel) | + LinearMask-GS | Reduction |
| *amsterdam* (Bungee-NeRF) | 27.8570 | 27.9720 | +0.1150 | 648,221 | 380,726 | 41.27% |
| *bicycle* (Mip-NeRF 360) | 25.1034 | 25.0053 | −0.0981 | 600,012 | 356,827 | 40.53% |
| *garden* (Mip-NeRF 360) | 27.1481 | 27.0226 | −0.1255 | 500,045 | 300,877 | 39.83% |
| **Mean (2 added scenes)** | 26.1258 | 26.0140 | −**0.1118** | 550,029 | 328,852 | **40.18**% |

Table 14: Decomposition of the mask-learning speed gap between 3DGS LinearMask and the FastGS-backbone variant (LinearMask-GS★) on Mip-NeRF 360, both measured on the same RTX 4090. Both per-factor speedups are directly measured. The #Gaussians values are averaged over the nine scenes. The two factors combine to $6.36 \times 3.7 \approx 24\times$, consistent with the measured mask-learning times of 110 s (3DGS backbone) and 4.6 s (FastGS backbone), since $110/24 \approx 4.6$.

| # | Factor | 3DGS LinearMask | FastGS (LinearMask-GS★) | Speedup |
|---|---|---|---|---|
| 1 | #Gaussians at mask window | Pre-prune 3.14 M (9-scene avg) | Pre-prune 0.493 M (9-scene avg) | 6.36× |
| 2 | Rasterizer implementation | Standard Inria rasterizer (full bbox per splat) | FastGS rasterizer (far fewer tiles per splat) | 3.7× |
| | | | **Combined speedup** ($6.36 \times 3.7$) | ≈**24**× |
| | | | Measured mask-learning time (3DGS vs. FastGS backbone) | 110 s vs. 4.6 s |
| | | | Consistency check: $110\,s/24 \approx 4.6\,s$ | ✓ |

Table 15: Net training-time effect of LinearMask-GS$^\star$ on the FastGS backbone, averaged over the nine Mip-NeRF 360 scenes. The added mask-learning cost (4.6 s) is more than offset by the faster 20k–30k iteration range (10.9 s saved from rendering fewer primitives), for a net saving of 6.3 s per scene.

| Component | Time (s) |
|---|---|
| 20k–30k range, FastGS native | 39.2 |
| 20k–30k range, LinearMask-GS$^\star$ | 28.3 |
| Range speedup | −10.9 |
| Added mask-learning cost | +4.6 |
| **Net change per scene** | **−6.3** |



\end{document}